\documentclass[11pt, a4paper, onecolumn, thu]{thuc3i}

\usepackage{amsmath,amsfonts,bm}

\def\eqref#1{equation~\ref{#1}}
\def\1{\bm{1}}

\DeclareMathAlphabet{\mathsfit}{\encodingdefault}{\sfdefault}{m}{sl}
\SetMathAlphabet{\mathsfit}{bold}{\encodingdefault}{\sfdefault}{bx}{n}

\usepackage{hyperref}
\usepackage{url}
\usepackage{graphicx} 
\usepackage{enumitem}
\usepackage{array}
\usepackage{wrapfig}
\usepackage{booktabs}
\usepackage{xcolor}
\usepackage{colortbl}
\usepackage{multirow}
\usepackage{adjustbox}
\usepackage{fontawesome5}
\usepackage{tcolorbox}
\usepackage{verbatim}
\usepackage{float}
\usepackage{graphicx}
\usepackage{amsmath} 
\usepackage{subfig}
\usepackage{tcolorbox}
\tcbuselibrary{skins,breakable}
\usepackage{fontawesome5}
\usepackage{marvosym}
\usepackage[authoryear, sort&compress, round]{natbib}

\definecolor{Gold}{rgb}{1, 0.88, 0.22}
\definecolor{Silver}{rgb}{0.87, 0.87, 0.87}
\definecolor{Bronze}{rgb}{0.88, 0.62, 0.40}

\colorlet{GoldD}{Gold!95!black}
\colorlet{SilverD}{Silver!95!black}
\colorlet{BronzeD}{Bronze!95!black}
\newcommand{\mc}[2]{\textbf{\textcolor{#1D}{#2}}}

\newcommand{\medalbox}[2]{\begingroup\setlength{\fboxsep}{0.5pt}\colorbox{#1}{\strut #2}\endgroup}

\setheadertext{P1-VL Technical Report}

 \title{P1-VL: Bridging Visual Perception and Scientific Reasoning in Physics Olympiads}
 \vspace{-5mm}
\author{
Yun Luo$^*$\textsuperscript{\Letter$\dagger$},
Futing Wang$^*$,
Qianjia Cheng$^*$,
Fangchen Yu$^*$,
Haodi Lei,
Jianhao Yan,
Chenxi Li,
Jiacheng Chen,
Yufeng Zhao,
Haiyuan Wan,
Yuchen Zhang,
Shenghe Zheng, 
Junchi Yao,
Qingyang Zhang,
Haonan He,
Wenxuan Zeng,
Li Sheng,
Chengxing Xie,
Yuxin Zuo,
Yizhuo Li,
Yulun Wu,
Rui Huang,
Dongzhan Zhou,
Kai Chen,
Yu Qiao,
Lei Bai\textsuperscript{\Letter},
Yu Cheng\textsuperscript{\Letter},
Ning Ding\textsuperscript{\Letter},
Bowen Zhou\textsuperscript{\Letter},
Peng Ye\textsuperscript{\Letter},
Ganqu Cui\textsuperscript{\Letter$\dagger$}\\
\vspace{1mm}
\centering{\normalsize P1 Team, Shanghai AI Laboratory}\\
\vspace{1mm}
$^*$ Equal Contribution~~  \textsuperscript{\Letter} Corresponding Authors~~ $^\dagger$ Technical Leads\\
\vspace{1mm}
\faEnvelope[regular]~\texttt{cuiganqu@pjlab.org.cn, luoyun1@pjlab.org.cn}  \quad
\faGithub~\href{https://prime-rl.github.io/P1-VL/}{P1-VL Tech Blog}
}
\vspace{-7mm}

\vspace{-3mm}
\begin{abstract}
The transition from symbolic manipulation to science-grade reasoning represents a pivotal frontier for Large Language Models (LLMs), with physics serving as the critical test anchor for binding abstract logic to physical reality. Physics demands that a model maintain physical consistency with the laws governing the universe, a task that fundamentally requires multimodal perception to ground abstract logic in reality. At the Olympiad level, diagrams are often constitutive rather than illustrative, containing essential constraints, such as boundary conditions and spatial symmetries, that are absent from the text. To bridge this visual-logical gap, we introduce P1-VL, a family of open-source vision-language models engineered for advanced scientific reasoning. Our method harmonizes \textit{Curriculum Reinforcement Learning}, which employs progressive difficulty expansion to stabilize post-training, with \textit{Agentic Augmentation}, enabling iterative self-verification at inference. Evaluated on HiPhO, a rigorous benchmark of 13 exams from 2024–2025, our flagship \texttt{P1-VL-235B-A22B} becomes the first open-source Vision-Language Model (VLM) to secure 12 gold medals
and achieves the state-of-the-art performance in the open-source models. Our agent-augmented system achieves the No.2 overall rank globally, trailing only Gemini-3-Pro. Beyond physics, P1-VL demonstrates remarkable scientific reasoning capacity and generalizability, establishing significant leads over base models in STEM benchmarks. By open-sourcing P1-VL, we provide a foundational step toward general-purpose physical intelligence to better align visual perceptions with abstract physical laws for machine scientific discovery.
\end{abstract}
\vspace{-3mm}
\begin{document}

\maketitle

\begin{figure}[h]
    \centering
    \includegraphics[width=0.92\linewidth]{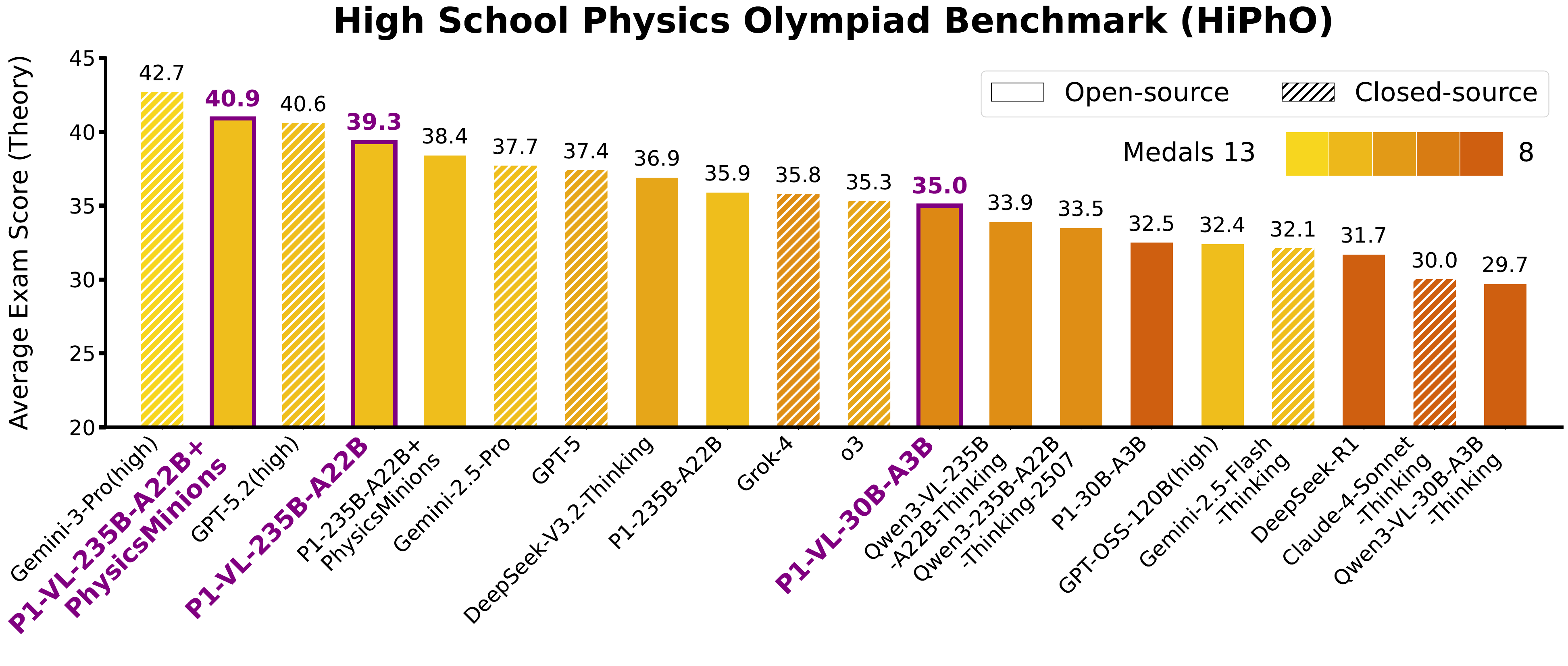}
    \vspace{-4mm}
    \caption{\texttt{P1-VL-235B-A22B} stands as the state-of-the-art open-source VLM in the Physics Olympiad benchmark (HiPhO), placing No.3 behind Gemini-3-Pro(high) and GPT-5.2(high) and achieving 12 gold medals. Even at mid-scale, \texttt{P1-VL-30B-A3B} achieved 9 gold medals, with a higher average score than most of the open-source models except P1-235B-A22B and DeepSeek-V3.2-Thinking. With the PhysicsMinions agent framework, \texttt{P1-VL-235B-A22B+PhysicsMinions} ranks No.2 on HiPhO.}
    \label{fig:first}
    \vspace{-4mm}
\end{figure}

\newpage
\begingroup
\setlength{\baselineskip}{1.1\baselineskip}
\tableofcontents
\endgroup
\newpage

\section{Introduction}
% Recent advances in Large Language Models (LLMs)~\citep{yang2025qwen3,Gemini-2.5} have propelled artificial intelligence beyond symbolic manipulation toward science-grade reasoning~\citep{gibney2025deepmind_sci,zhang2024llmsci_survey,Intern-S1}. Among scientific disciplines, \textbf{physics} stands as the critical crucible for these emerging capabilities. Unlike mathematical reasoning tasks, physics demands a rigorous synthesis of abstract laws and causal logic to model the behavior of the universe. Consequently, the ability to solve complex physics problems serves as a critical benchmark, distinguishing models that merely retrieve information from those capable of genuine, first-principles reasoning.

Recent advances in Large Language Models (LLMs)~\citep{yang2025qwen3,Gemini-2.5} have propelled artificial intelligence beyond symbolic manipulation toward science-grade reasoning~\citep{gibney2025deepmind_sci,zhang2024llmsci_survey,Intern-S1}. Among scientific disciplines, \textbf{physics} stands as the critical test anchor for these emerging capabilities. Physics inherently demands a rigorous synthesis of {multimodal perception} and causal logic to model the behavior of the universe. Consequently, the ability to solve complex physics problems, a task that often requires mapping physical intuition to abstract laws, provides a rigorous test for genuine, first-principles reasoning.

\begin{figure}[tph]
    \centering
    \includegraphics[width=0.98\linewidth]{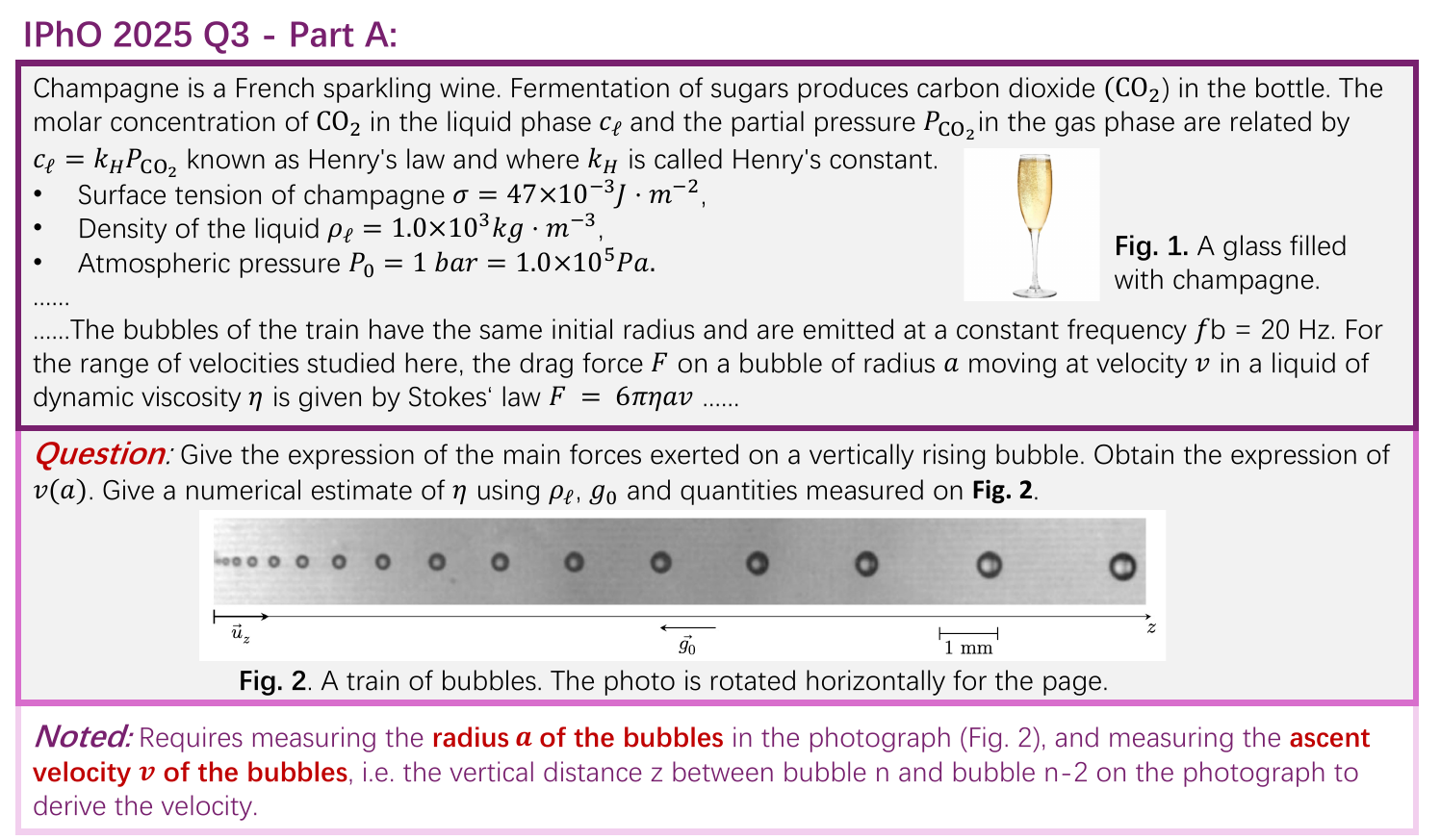}
    \vspace{-1.5mm}
    \caption{A question sample from the International Physics Olympiad 2025 (IPhO 2025), where the question requires measuring the radius of bubbles and estimating their velocity in Fig 2. }
    \label{fig:data_example}
\end{figure}

% This reasoning capability is most rigorously tested in 
\textbf{Olympiad-level competitions}, such as the International Physics Olympiad (IPhO) serve as a significant surrogate to evaluate the physics reasoning capacity of language models, where problems demand deep conceptual understanding and precise system decomposition. While recent works have begun to adapt LLMs for these challenges~\citep{qiu2025physicsagent,physicsminions,chen2025p1masteringphysicsolympiads}, they predominantly focus on textual reasoning, overlooking a fundamental characteristic of physical problem-solving: its inherent multimodality. In many Olympiad problems, diagrams are not merely illustrative but constitutive—they contain essential geometric constraints, circuit topologies, and force interactions that are purposefully omitted from the text. Without the ability to perceive and interpret these visual schematics, a text-only model faces an insurmountable information gap. As illustrated in Figure \ref{fig:data_example} (IPhO 2025), the task involves quantifying bubble radii and ascent velocities directly from visual inputs, effectively simulating the data analysis process in real-world physics experiments.
Thus, Physics Olympiads serve as high-fidelity testbeds for Vision-Language Models (VLMs), verifying the critical ability to align visual observations with abstract physical laws~\citep{2025hipho}.

\vspace{-2.5mm}
    \vspace{-1.5mm}

Successfully bridging this visual-logical gap is more than an academic achievement; it lays the foundation for \textbf{general-purpose physical intelligence}. We posit that mastering the rigorous constraints of Olympiad problems is a necessary precursor to achieving \emph{machine scientific discovery}~\citep{zheng2025llm-discover} and reliable \emph{embodied AI}. Just as a scientist needs a theory and a robot needs a world model, future AI systems must first internalize the laws of physics in a controlled environment. Therefore, our focus on Olympiad-level reasoning is not an end in itself, but a strategic step toward building systems capable of understanding and eventually navigating the physical reality.

In this work, we introduce \textbf{P1-VL}, a new family of open-source vision-language models designed to advance the frontier of scientific AI. Our framework harmonizes two critical paradigms: \emph{Curriculum Reinforcement  Learning (RL) Training} and \emph{Agentic Augmentation}. This dual approach ensures that the model not only internalizes robust reasoning capabilities through RL~\citep{sutton1998rl-basis}, but also effectively operationalizes them via adaptive agentic control during inference~\citep{internagent1.5}.
\vspace{-2mm}
\begin{itemize}[left=0pt]
    \item \textbf{Curriculum RL Training.} 
    P1-VL is developed exclusively through RL post-training~\citep{guo2025deepseek-r1,cui2025prime-rl} built upon base vision-language models. To maximize the efficacy of this paradigm, we introduce a  {curriculum RL framework} that systematically advances reasoning capabilities through progressive difficulty expansion, and robust stabilization mechanisms. By ensuring a structured learning trajectory, our design enables sustained, long-term optimization and effectively circumvents notorious RL pitfalls such as reward sparsity, entropy collapse, and training stagnation.
    \vspace{1mm}
    \item \textbf{Agentic Augmentation.} 
    During the inference phase, we integrate P1-VL models with the \textit{PhysicsMinions} agent framework~\citep{physicsminions} to enable iterative correction and self-verification. This mechanism facilitates multi-turn reflection, empowering the model to reason, critique, and refine its solutions—mirroring the rigorous cognitive process of human physicists. By leveraging this structured test-time reasoning, P1-VL effectively extends its problem-solving depth and reliability without requiring additional training parameters.
\end{itemize}
\vspace{-2mm}
    
We release two variants of the P1-VL family and evaluate them on {HiPhO}~\citep{2025hipho}, a rigorous benchmark aggregating 13 recent Olympiad exams from 2024–2025. Our flagship model, \texttt{P1-VL-235B-A22B}, marks a significant milestone for the open-source community: it is the first specialized physics vision-language model to secure \textbf{12 golds and 1 silver} across the full HiPhO suite. Remarkably, the standalone P1-VL surpasses even the agent-enhanced text baseline (\texttt{P1-235B-A22B+PhysicsMinions} \citep{chen2025p1masteringphysicsolympiads}), ranking \textbf{3rd} among all open and closed-source models. Meanwhile, our lightweight variant, \texttt{P1-VL-30B-A3B}, achieves 9 gold medals, surpassing the average score of most open-source models. When further augmented with the \textit{PhysicsMinions} agent framework, P1-VL ascends to the \textbf{2nd overall spot} on the leaderboard.

Beyond its specialized domain, P1-VL demonstrates remarkable {generalizability}.
In the multidisciplinary FrontierScience-Olympiad, both \texttt{P1-VL-235B-A22B} and \texttt{P1-VL-30B-A3B} models demonstrate robust gains across biology, chemistry, and physics, outperforming their respective baselines by margins of 8.0 and 9.1 points, respectively.
In the mathematical reasoning tasks and multi-modal STEM reasoning tasks, our model could also consistently outperform their base models.
This superiority extends to out-of-domain evaluations, where both variants consistently eclipse their base models across diverse text, and multi-modal benchmarks.

\vspace{-4mm}
\paragraph{Contributions.} 
This work makes the following key contributions:
\vspace{-3mm}
\begin{itemize}[left=0pt]
    \item We introduce \textbf{P1-VL}, a first family of open-source Vision-Language Model specialized in physics problems, achieving 12 gold medals in physics Olympiad competitions. 
    \item We develop a \textbf{curriculum RL framework} with difficulty expansion, and training stabilization for sustained reasoning improvement.
    \item We open-source the P1-VL model family, filling the multimodal void within the P1 ecosystem \citep{chen2025p1masteringphysicsolympiads}, and bridging the gap between visual perception and scientific reasoning.
\end{itemize}
Collectively, the progress lays the foundation for AI systems that not only interpret reality but eventually contribute to the frontiers of physics research and the development of robust world models.

% \section{Related Works (optional)}
\section{Physics Dataset}

\subsection{Overview}

\begin{figure}[t]
    \centering 
    \begin{minipage}[c]{0.56\linewidth}
        \centering
        \small 
        \begin{tabular}{l r}
            \toprule
            \textbf{Statistics} & \textbf{Number} \\
            \midrule
            \multicolumn{2}{l}{\textit{Data Composition}} \\
            Total Problems & 8,033 \\
            \quad - Problems from Olympiads & 4,126 (51\%) \\
            \quad - Problems from textbooks & 3,907 (49\%) \\
            Total Answers & 10,516\\
            Total Fields & 5 \\
            Total Subfields & 25 \\
            Total Answer Types & 5 \\
            \midrule
            \multicolumn{2}{l}{\textit{Data Sources}} \\
            Textbooks & 20 \\
            Olympiad Types & 9 \\
            \quad - Sets Collected & 199 \\
            \midrule
            \multicolumn{2}{l}{\textit{Token Statistics}} \\
            Average Question Tokens & 943 \\
            Max Question Tokens & 6,604 \\
            Max Solution Tokens & 5,519 \\
            \midrule
            \multicolumn{2}{l}{\textit{Images Statistics}} \\
            Questions with Images & 5,513 \\
            Questions without Images & 2,520\\
            Max Images per Question & 8\\
            Average Images per Question & 0.82 \\
            \bottomrule
        \end{tabular}
        \captionof{table}{Statistics of the multi-modal training data.}
        \label{tab:data_statistics}
    \end{minipage}
    \hfill 
    \begin{minipage}[c]{0.43\linewidth}
        \centering
        \includegraphics[width=\linewidth]{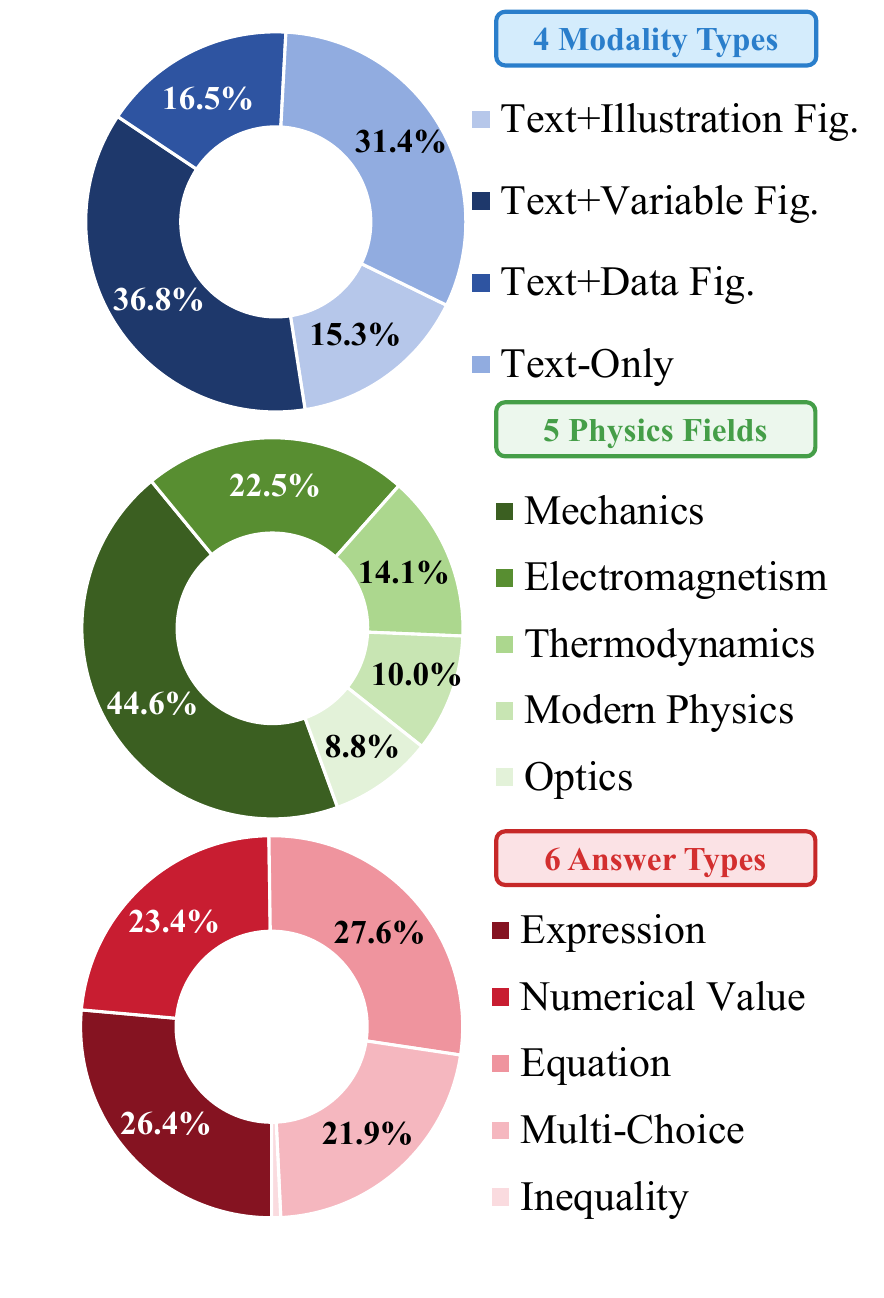}
        \caption{Distribution of the training data.}
        \label{fig:field_distribution}
    \end{minipage}

\end{figure}

We introduce a systematically curated multimodal dataset of 8,033 problems designed to advance genuine scientific reasoning in VLMs. The dataset includes problems from physics Olympiads (4,126), undergraduate textbooks (2,968), and competition guides (939), as summarized in Table~\ref{tab:data_statistics}. 
% Rather than focusing on scale, we prioritize {multimodal richness}: a significant proportion of the data includes essential diagrams, demanding the models to perceive the physics variables and data.
% Following protocols used in PHYSICS~\citep{zheng2025scaling} and HiPhO~\citep{2025hipho}, we construct the annotation pipeline by integrating human verification with model-assisted extraction to ensure higher data fidelity. 

\begin{figure}
    \centering
    \includegraphics[width=0.99\linewidth]{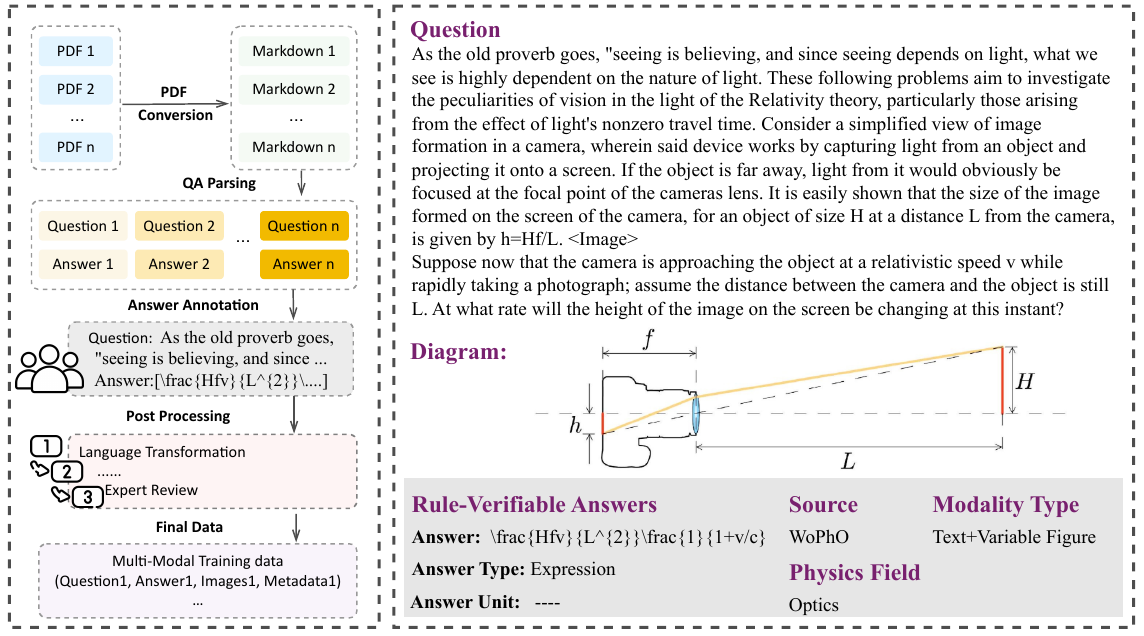}
    \caption{Data collection pipeline for physics data.}
    \label{fig:placeholder}
\end{figure}

\subsection{Dataset Construction}
\textbf{Data Collection.}
Our data construction is driven by a core objective: enabling VLMs to internalize structured reasoning aligned with physical laws and empirical consistency—a prerequisite for genuine scientific intelligence. To achieve this, we prioritize problems that combine conceptual depth with rule-verifiable outcomes, traits shown to catalyze reasoning improvements~\citep{guo2025deepseek-r1,yang2025qwen3,wen2025reinforcement,chen2025p1masteringphysicsolympiads}. Within the physics domain, physics Olympiad problems naturally contain such properties: they demand rigorous modeling and multi-step inference while remaining formally tractable. The value of this specific difficulty profile is underscored by empirical findings in PHYSICS~\citep{zheng2025scaling}, which demonstrate that LLMs face greater challenges in Olympiad tasks than in introductory undergraduate problems. Accordingly, we first curate our dataset from two complementary pillars: ten prestigious physics Olympiads (up to 2023, including IPhO and APhO) that capture a graded difficulty spectrum, and ten authoritative competition textbooks that provide systematically organized exercises with expert-verified solutions. To further augment the dataset's multi-modal richness, we extend our collection to include undergraduate textbooks. These sources serve as a vital repository of image-centric problems, significantly enhancing the overall density of visual schematics and diagrams within the corpus.

\textbf{Data Annotation.}
Driven both by the demand for high-quality extraction and the heterogeneity of the sources, the construction process is organized as a multi-stage pipeline (Figure \ref{fig:placeholder}) following the pipeline of P1\citep{chen2025p1masteringphysicsolympiads}.
% \textbf{(1) PDF-to-Markdown Conversion.} Source materials in PDF format are parsed into Markdown using Optical Character Recognition (OCR) tools. 
%      \textbf{(2) Questions and Solutions Parsing.} Extraction strategies are tailored to the two source types. For textbooks, model-assisted parsing leverages structural cues (e.g., chapter boundaries and numbering) to align exercises with their solutions. Olympiad problems, characterized by lengthy statements and multiple sub-questions, are manually restructured by experts to separate shared background from sub-questions, preserving both readability and fidelity. 
%      \textbf{(3) Answer Annotation.} Answers are automatically extracted by models and decomposed into structured lists, allowing each sub-answer to be individually validated against model outputs. Units are separated into explicit fields to support standardized, rule-based scoring.
%      \textbf{(4) Language Transformation.} Problems originating in Chinese (e.g., CPhO) are translated into English with Claude-3.7-Sonnet to enhance the data quantity.

\subsection{Quality Control.} 
To ensure data reliability, we implement a rigorous multi-stage pipeline integrating automated model-based filtering with human verification. \textbf{(1) OCR Correction.} Texts parsed from complex layouts or low-quality scans are manually validated against original PDFs to rectify OCR artifacts. \textbf{(2) Answer Cross-Validation.} Three models --- Gemini-2.5-Flash, Claude-3.7-Sonnet, and GPT-4o --- independently extract answers from each question–solution pair. Consensus is established via majority vote (at least two models agree); non-consensus items are discarded. \textbf{(3) Task Filtering.} Problems requiring diagram generation or involving unverifiable open-ended answers (e.g., proofs) are removed. \textbf{(4) Visual Consistency Check.} We utilize Gemini-2.5-Flash to detect and filter out samples where the question text references a figure that is missing from the input. \textbf{(5) Expert Review.} A final consistency check is performed by Claude-3.7-Sonnet, followed by targeted manual refinement by experts. Through this filtering process, the dataset is refined from 13,432 to 8,033 items, yielding a high-fidelity, multi-modal bilingual corpus well-suited for RLVR.

\section{Approach}
\subsection{RL Formulation}

We formulate the challenge of solving Physics Olympiad problems through the lens of reinforcement learning (RL)~\citep{sutton1998rl-basis} process. 
Formally, let $\mathcal{M} = (\mathcal{S},  \mathcal{A}, P, r)$ denote the underlying Markov Decision Process (MDP), where:
\begin{itemize}[left=0pt]
    \item $\mathcal{S}$ represents the state space, corresponding to the model context, including the problem statement and all previously generated reasoning tokens.
    \item $\mathcal{A}$ is the action space, which constitutes the discrete action space of possible tokens in the vocabulary.
    \item $P(s' \mid s, a)$ defines a (deterministic) state transition function, which appends the newly chosen action $a$ to the state $s$, resulting in an updated context $s'$.
    \item $r(s, a)$ provides a scalar reward indicating the correctness and quality of the finalized reasoning trajectory.
\end{itemize}
The learning objective is to optimize the policy $\pi _{\theta}$ by maximizing the expected return:
\begin{equation}
    J(\pi_\theta) = \mathbb{E}_{\tau \sim \pi_\theta}\Bigg[\sum_{t=0}^{T} r(s_t, a_t)\Bigg],
\end{equation}
where $\pi_\theta(a_t|s_t)$ is the policy parameterized by model parameters $\theta$, and $\tau = (s_0, a_0, \dots, s_T)$ denotes a trajectory sampled from $\pi_\theta$.

\textbf{Policy Gradient.}  
The policy gradient~\citep{sutton1999pg-sutton} method optimizes $\pi_\theta$ by ascending the gradient of the expected return:
\begin{equation}\label{eq:pg_loss}
    \nabla_\theta J(\pi_\theta) = \mathbb{E}_{\tau \sim \pi_\theta}\Bigg[\sum_{t=0}^{T} \nabla_\theta \log \pi_\theta(a_t \mid s_t) \, A^\pi(s_t, a_t)\Bigg],
\end{equation}
where $A^\pi(s_t,a_t)$ is the advantage function estimating the relative value of action $a_t$ in state $s_t$.  
We adopt this standard form and instantiate it via GSPO below.

\textbf{Group Sequence Policy Optimization (GSPO).} 
GSPO~\citep{zheng2025gspo} elevates optimization from the token level~\citep{shao2024grpo,yu2025dapo} to the sequence level, employing length-normalized sequence likelihood importance ratios:
\begin{equation}
    \rho_i(\theta) = \left(\frac{\pi_\theta(y_i|x)}{\pi_{\theta_{\text{old}}}(y_i|x)}\right)^{1/|y_i|} = \exp\left(\frac{1}{|y_i|}\sum_{t=1}^{|y_i|} \log \frac{\pi_\theta(y_{i,t}|x, y_{i,<t})}{\pi_{\theta_{\text{old}}}(y_{i,t}|x, y_{i,<t})}\right),
\end{equation}
where $|y_i|$ denotes the sequence length, and the $1/|y_i|$ term implements length normalization to reduce variance. The corresponding advantage function is computed at the sequence level:
\begin{equation}
    \hat{A}_i^{\text{GSPO}} =\frac{R_i - \text{mean}(\{R_j\}_{i=1} ^{G}) }{\text{std}(\{R_j\}_{i=1} ^{G})},
\end{equation}
with the objective function:
\begin{equation}
    J^{\text{GSPO}}(\theta) = \mathbb{E}_{x \sim \mathcal{D}, \{y_i\}_{i=1}^G \sim \pi_{\theta_{\text{old}}}(\cdot|x)}\left[\frac{1}{G}\sum_{i=1}^G \min\left({\rho}_i(\theta)\hat{A}_i^{\text{GSPO}}, \text{clip}(s_i(\theta), 1-\epsilon, 1+\epsilon)\hat{A}_i^{\text{GSPO}}\right)\right].
\end{equation}

\subsection{Instantiation}
We keep the same design of the reward and the verifier following P1 Series \citep{chen2025p1masteringphysicsolympiads}.

\textbf{Reward Design. } We employ a binary reward scheme based on answer correctness, leveraging the fact that our physics dataset contains verifiable ground-truth outcomes:
\begin{equation}
    r = \begin{cases}
        1, & \text{if the predicted answer matches the ground truth}, \\
        0, & \text{otherwise}.
    \end{cases}
\end{equation}
% However, physics problems often involve multiple sub-questions or require multiple final results (e.g., solving for both $a$ and $b$).  
The we adopt the test-case-style reward aggregation, defining the final reward as:
\begin{equation}
    R = \frac{1}{N}\sum_{i=1}^N r_i,
\end{equation}
where $N$ is the number of required sub-answers in the problem, and $r_i$ denotes the correctness indicator for the $i$-th sub-answer. Furthermore, we apply the system prompt as shown in Figure~\ref{fig:system_prompt} to extract the multi-box answers. 

\begin{figure}[h]
\begin{tcolorbox}[colback=gray!5!white, colframe=gray!75!black, title=System Prompt of Multi-box Style]
\begin{verbatim}
Please answer the problem adhering to the following rules:
1. Please use LaTeX format to represent the variables and formulas 
   used in the solution process and results.
2. Please put the final answer(s) in \boxed{}, note that the unit 
   of the answer should not be included in \boxed{}.
3. If the problem requires multiple answers, list them in order, 
   each in a separate \boxed{}.
\end{verbatim}
\end{tcolorbox}
\caption{System prompt design for P1-VL training.}
\label{fig:system_prompt}
\vspace{-2mm}

\end{figure}

\textbf{Verifier Design.}
Physics problems frequently necessitate solutions in the form of complex symbolic expressions rather than scalar values, posing a challenge for standard verification metrics. We implement the hybrid verification framework that harmonizes rigorous symbolic logic with semantic model evaluation: (1) Rule-Based Verifier, we integrate symbolic computation libraries (SymPy~\citep{meurer2017sympy}) with heuristic strategies from math-verify~\citep{math-verify}; (2) To capture correct answers that elude strict symbolic matching, we adopt the XVerify~\citep{chen2025xverify} paradigm. We employ a specialized LLM (Qwen3-30B-A3B-Instruct) as a semantic judge.

\subsection{Technical Design}

\subsubsection{Curriculum Training}
\textbf{Preliminary Difficulty Calculation and Filtering Strategy.} The distribution of data difficulty and data quality plays a pivotal role in the convergence of RL training \citep{zhang2025right,cui2025prime-rl}. To construct an effective training set, we adopt an inference-based metric to estimate the solvability of each instance. Let $D(x_i, y_i)$ denote the empirical pass rate:
\begin{equation}
\label{eq:difficulty}
D(x_i,y_i) = \frac{1}{N}\sum_{k=1}^N \mathbb{I}(\hat{y}_k = y_i),
\end{equation}
where $\hat{y}_k$ denotes the $k$-th response generated by the model. We utilize the Qwen3-VL-30B-A3B model to perform extensive rollouts (with $N=72$) on the training dataset. The correctness of inference rollouts is determined only by the rule-based verifier. Based on the distribution of $D(x_i, y_i)$, we apply a dual-end filtering strategy:

\vspace{-2mm}
\begin{enumerate}[left=0pt]
    \item \textbf{Pruning Trivial Samples ($D(x_i, y_i) > 0.7$): } Samples that the model can already solve with high confidence contribute minimal gradient variance during optimization, accelerating entropy collapse of the policy model \citep{cui2025entropy}. To improve training efficiency, we categorize samples with $D > 0.7$ as "trivial" and remove them from the training set following \cite{chen2025p1masteringphysicsolympiads}.
    % This ensures the model resources are allocated to borderline cases where improvement is most needed. 
    \vspace{2mm}
    \item \textbf{Recovering Zero-Shot Failures ($D(x_i, y_i) = 0.0$):} Samples where the pass rate is strictly $0.0$ present a challenge; they represent either hard-core reasoning steps or malformed questions. Training on the malformed questions can lead to reward hacking or instability. Thus, we filter these malformed samples and introduce a \textit{recovery pipeline} using Gemini-2.5-Flash. We prompt such a stronger model to (a) verify the alignment between the question and image, and (b) verify the completeness of the question, including physical factors or numerical values, (c) refine ambiguous problem descriptions. The unrecoverable tasks are finally excluded to maintain the overall quality. This process converts previously "unsolvable" noise into effective, clear RL tasks.
\end{enumerate}

\textbf{Curriculum RL Training.}  
% Even with high-quality data, training stability is often compromised by the low success rates inherent in difficult scientific problems. To address this, we implement {difficulty expansion}, a curriculum that initiates training on tractable samples to stabilize policy gradients before graduating to complex tasks. 
% \vspace{-2mm}
% \begin{enumerate}[left=0pt]
     % \textbf{Difficulty expansion.} 
     Directly training with RL on difficult problems for the VLMs could cause unstable policy gradients and low efficiency because of the low success rates, thus, we expand the difficulty of the tasks step by step. In the initial stage, we train the model with samples that exhibit a pass rate below an upper threshold of $0.7$. By filtering out trivial instances (i.e., those with success rates $D(x_i, y_i) > 0.7$) early in the process, we prevent the model from wasting computation on tasks it has already mastered. Subsequently, we further intensify the training challenge by lowering this inclusion threshold. This progressive tightening of the difficulty constraint effectively prunes moderately easy samples from the dataset, compelling the model to concentrate its optimization capacity exclusively on the harder tail of the distribution, where its performance is still lacking.

     As the curriculum shifts toward the harder tail of the distribution, static rollout configurations become restrictive, failing to provide the search depth required for these escalated challenges~\citep{cui2025entropy,deepscaler2025}. Thus, we follow P1\citep{chen2025p1masteringphysicsolympiads} to dynamically expand the exploration space, such as group size and generation window, in pace with the difficulty expansion, ensuring the search budget scales adequately to sustain long-term learnability.
    % \vspace{2mm}
    
    %  \textbf{Group size and generation window expansion.}  
    % We progressively scale both the breadth and depth of the model's exploration during curriculum training.
    % In the GSPO~\citep{zheng2025gspo} group-based advantage estimation framework, each training query $q$ is associated with a group of $G$ sampled responses $\{o_i\}_{i=1}^G \sim \pi_{\theta_{\text{old}}}(\cdot|q)$to estimate the advantages. By increasing the group size $G$, we enhance the probability of capturing rare, high-quality trajectories that are essential for solving complex problems with low intrinsic success rates.  Simultaneously, we extend the maximum generation window to prevent reasoning chains from being prematurely truncated. This dual-expansion strategy allows the model to encounter more informative learning signals and effectively learn from them.
    % \vspace{2mm}
    % \item \textbf{Generation window expansion.} 
    % The maximum output length (generation window) imposes a constraint on the depth of reasoning the model can express. When this window is restricted, reasoning chains for complex physics problems are susceptible to premature truncation, resulting in incomplete or incorrect answers. To address this, we progressively extend the generation window throughout the training process. This strategy allows the model to develop longer and more coherent reasoning chains, thereby enhancing its capacity to solve high-complexity problems and mitigating errors caused by truncation.
% \end{enumerate}
% \vspace{-2mm}

\subsubsection{Training Stabilization Mechanism}

\paragraph{Mitigate Train-inference Mismatch}
Recent studies~\citep{yao2025tis,jiacai2025speed} have noticed that the train-inference engine difference is a key cause of instability in RL training.
An empirical analysis is shown in Section \ref{sec:tim}. 
To formally understand this statement, Eq.~\ref{eq:pg_loss} could be rewritten as,
\begin{equation}
    \nabla_\theta J(\pi_\theta) = \mathbb{E}_{\tau \sim \pi_\theta^{rollout}}\Bigg[\sum_{t=0}^{T} \nabla_\theta \log \pi_\theta^{train}(a_t \mid s_t) \, A^\pi(s_t, a_t)\Bigg],
\end{equation}
where $\pi_\theta^{rollout}$ denotes the policy used to generate trajectories during rollout, and $\pi_\theta^{train}$ denotes the policy evaluated during gradient computation.  RL frameworks~\citep{slime_github,sheng2024verl} often adopt different engines for rollout (e.g. vllm~\citep{kwon2023vllm} and SGLang~\citep{zheng2024sglang}) and 
training (e.g. FSDP~\citep{merry2021fsdp} and Megatron~\citep{shoeybi2019megatron}), which 
% While this design significantly improves throughout, it inadvertently 
introduce a mismatch between $\pi_\theta^{rollout}$ and $\pi_\theta^{train}$ due to differences in numerical precision, computation optimization strategies, and
kernel implementations, leading to biased gradient estimates and training instability: 
$
    \pi_\theta^{rollout}(a|s) \neq \pi_\theta^{train}(a|s).
$

\textbf{Masked Importance Sampling}. To mitigate the instability caused by the training-inference mismatch, we adopt Sequence-level Masked Importance Sampling (Seq-MIS) \citep{jiacai2025speed}. Unlike token-level clipping methods, 
% which may introduce bias by ignoring state distribution shifts,
Seq-MIS employs a "hard trust region" by rejecting entire out-of-distribution trajectories. However, the cumulative importance weight $\rho(\tau)$ scales exponentially with sequence length $T$, making it difficult to set a consistent rejection threshold $C$ for tasks with varying lengths. To address this, we adopt the Geometric Mean of the importance weights to normalize the metric across different lengths. The gradient estimator is formulated as:
\begin{equation}
    \label{eq:geo_mis_gradient}
    \nabla_{\theta} J_{\text{Geo-MIS}}(\theta) = \mathbb{E}_{\tau \sim \pi_{\text{old}}} \left[ \underbrace{\mathbb{I}\left( \rho(\tau)^{1/T} \leq C \right)}_{\text{Geo-Mask}} \cdot \rho(\tau) \cdot \sum_{t=0}^{T} \nabla_{\theta} \log \pi_{\theta}(a_t | s_t) A^{\pi}(s_t, a_t) \right],
\end{equation}
where $\tau = (s_0, a_0, \dots, s_T, a_T)$ represents the sampled trajectory, and $T$ is the sequence length. The sequence-level importance weight is defined as $\rho(\tau) = \prod_{t=0}^{T} \frac{\pi^{\text{trainer}}_\theta(a_t|s_t)}{\pi^{\text{rollout}}_\theta(a_t|s_t)}$. 
$\mathbb{I}(\cdot)$ denotes the indicator function, rejecting trajectories whose geometric mean of importance weights exceeds the threshold $C$.
% By using the geometric mean $\rho(\tau)^{1/T}$, we establish a length-invariant trust region, effectively filtering out OOD samples based on their average per-token divergence while maintaining an unbiased estimator within the trust region.

\subsection{Training Dynamics}\label{train_dyn}

\textbf{Implementation.} For the implementation of P1-VL training pipeline, we adopt the VERL \citep{sheng2024verl} framework, which is an efficient VLMs post-training framework offering the RL implementation in Megatron.  We adopt \texttt{Qwen3-VL-30B-A3B-Thinking} and \texttt{Qwen3-VL-235B-A22B-Thinking} as starting points of models. To adaptively adjust the aforementioned learnability during training, we periodically resume training from the previous checkpoint with updated configurations. 

\textbf{Image Processing}
For questions containing images, the special token <image> is inserted at the corresponding position within the text. To facilitate unified RL training across modalities, we standardize the input format by padding all text-only samples with a blank image of resolution $448 \times 448$, and appending the <image> token to the end of the question. Vision-language models (VLMs), such as Qwen3-VL \citep{bai2025qwen3vltechnicalreport} and Intern3.5-VL \citep{wang2025internvl35advancingopensourcemultimodal}, typically consist of an image encoder, a language model, and a projection layer that bridges the two. Since our training data is not specifically curated for image alignment, we freeze the parameters of both the vision encoder and the projection layer during RL training. The analysis of training data composition is shown in Section \ref{sec:composition}. 
Furthermore, we empirically observe that padding images has a negligible impact on text-only performance. Therefore, we omit blank images during evaluation.

\textbf{Settings of Curriculum Training Phrases.}
We detail the stage-wise hyperparameter settings of the P1-VL post-training process in Table~\ref{tab:training_params}. Following our curriculum framework, we progressively amplify the training complexity by scaling the data difficulty (metric defined in Eq. \ref{eq:difficulty}) and expanding the exploration space via larger group sizes and extended generation windows. To bridge the train-inference gap, we consistently apply \textit{Masked Importance Sampling}—a correction technique for off-policy stability—across all stages. Notably, our optimization backbone is built upon the {GSPO} algorithm to ensure stable convergence for MoE architectures.  Furthermore, we exclusively employ rule-based verifiers during training to mitigate reward hacking.
% following \cite{chen2025p1masteringphysicsolympiads}.

\begin{table}[thbp]
\centering
\caption{Configuration of different phrases in P1-VL training.}
\label{tab:training_params}
\resizebox{0.8\linewidth}{!}{
\begin{tabular}{lccc|ccc}
\toprule
\multirow{2}{*}{\textbf{Model}} & \multicolumn{3}{c|}{\textbf{P1-VL-30B-A3B}} & \multicolumn{3}{c}{\textbf{P1-VL-235B-A22B}} \\
\cmidrule(lr){2-4} \cmidrule(lr){5-7}
& Stage 1 & Stage 2 & Stage 3 & Stage 1 & Stage 2 & Stage 3 \\
\midrule
\textbf{Group size} & 8 & 8 & 16 & 8 & 8 & 16  \\
\textbf{Difficulty} &(0.0,0.7]&(0.0,0.5]&[0.0,0.5]&(0.0,0.7]&(0.0,0.5]&[0.0.0.5] \\
\textbf{Generation Window} & 40k & 60k & 80k & 48k & 54k & 54k \\
\textbf{Learning Rate} & \multicolumn{3}{c|}{$1 \times 10^{-6}$} & \multicolumn{3}{c}{$1 \times 10^{-6}$} \\
\textbf{Algorithm} & \multicolumn{3}{c|}{GSPO w. MIS} & \multicolumn{3}{c}{GSPO w. MIS} \\
\textbf{Verifier Choice} & \multicolumn{3}{c|}{Rule-based Verifier Only} & \multicolumn{3}{c}{Rule-based Verifier Only} \\
\textbf{Rollout Batch Size} & \multicolumn{3}{c|}{2048} & \multicolumn{3}{c}{1024} \\
\textbf{Update Batch Size} & \multicolumn{3}{c|}{256} & \multicolumn{3}{c}{256} \\
\bottomrule
\end{tabular}
}
\end{table}

\begin{figure}[h]
    \centering
    % \includegraphics[width=\linewidth]{figs/data_example.pdf}
    % \vspace{-1mm}
    \includegraphics[width=\linewidth]{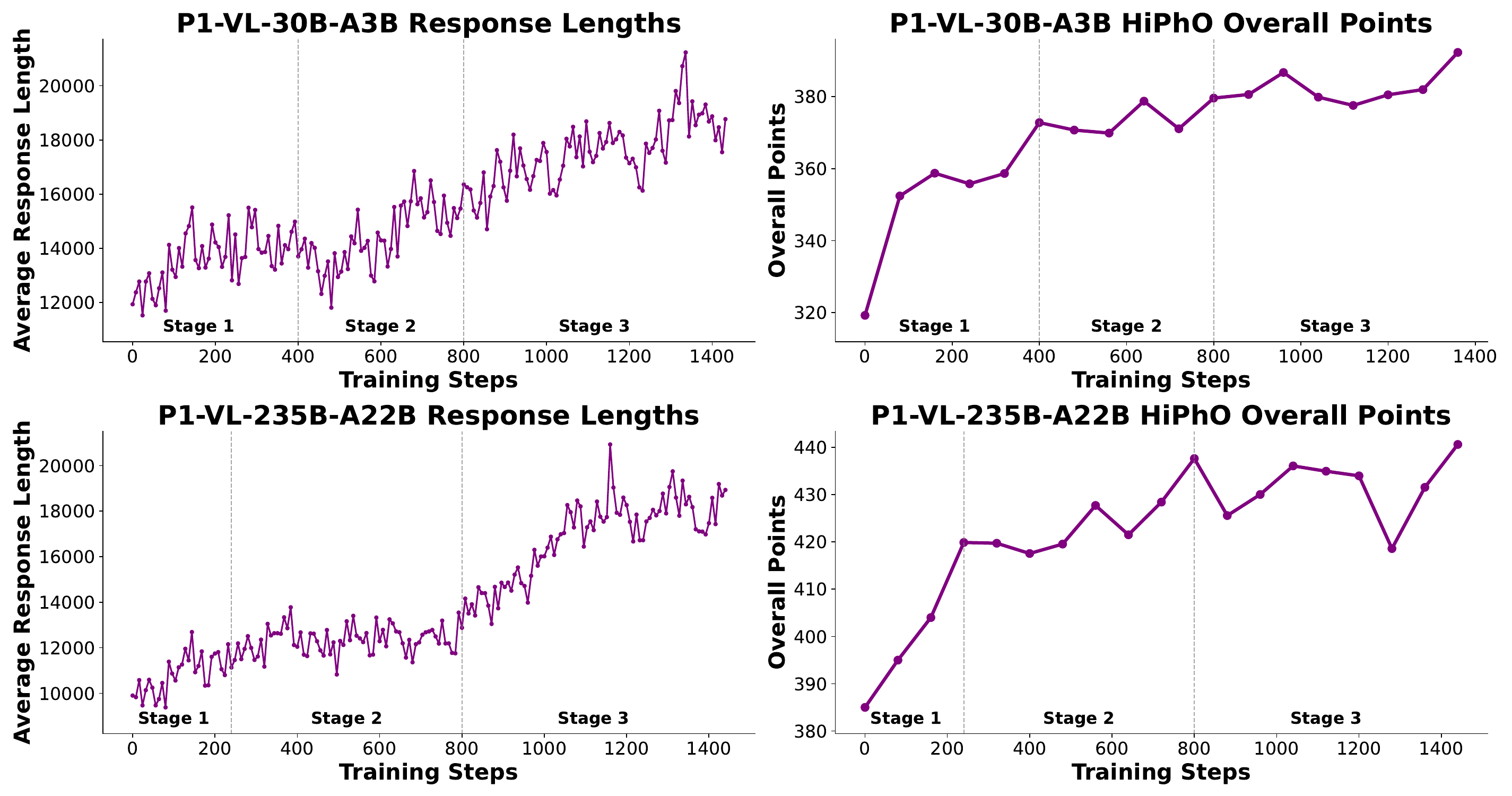}
    \caption{The training dynamics of P1-VL curriculum RL. The upper side shows the dynamics of P1-VL-30B-A3B, and the lower shows those of P1-VL-235B-A22B. The training is set in three stages with curriculum difficulty expansion, group size expansion, and generation window expansion. The average response length represents the generation length during training, and the overall points refer to the sum score of all the exams in HiPhO with model-based verifier (Qwen-3-30B-Instruct-2507).}
    \label{fig:dynamics}
    \vspace{-1mm}
\end{figure}

\textbf{Analysis.}
Figure~\ref{fig:dynamics} illustrates the training dynamics across the curriculum RL process. A key observation is the gradual increase in response length, indicating that the model is developing the capacity for deeper reasoning to tackle complex problems. To accommodate this evolving capability, we strategically expand the difficulty and meanwhile expand exploration space by increasing the generation window and group size at each stage transition. Quantitatively, validation results on HiPhO demonstrate a steady performance improvement throughout training, underscoring both the effectiveness and the stability of our algorithm. A detailed analysis of the effect of expansion is shown in Section \ref{sec:curriculum}.

\subsection{Agentic Augmentation}

At inference time, we enhance test-time reasoning with a unified agentic augmentation framework, \texttt{PhysicsMinions}~\citep{physicsminions}, a pioneering coevolutionary multi-agent system for multimodal scientific problem solving. The system is organized into three tightly coupled studios: the \emph{Visual Studio}, the \emph{Logic Studio}, and the \emph{Review Studio}. The Visual Studio performs structured perception and converts raw visual evidence into symbolic representations. The Logic Studio generates and iteratively refines candidate solutions through solver and introspector collaboration. The Review Studio validates solutions using both domain-specific and general verification. Together, these studios form a closed critique and refinement loop that enables scalable test-time reasoning and improves robustness on complex scientific tasks.

A key distinction from the earlier text-only P1 deployment of \texttt{PhysicsMinions}~\citep{chen2025p1masteringphysicsolympiads} is that P1-VL fully activates the Visual Studio to process diagrams, plots, and other visual inputs. By grounding reasoning in structured visual representations, the system can directly incorporate visual evidence into downstream symbolic reasoning, enabling effective handling of multimodal Olympiad-style problems.

Beyond multimodal perception, we extend the agent pipeline with a domain-adaptive mechanism that generalizes the framework to heterogeneous scientific disciplines. The system automatically selects discipline-specific solver prompts and domain verifiers according to the detected problem type, such as physics, chemistry, or biology, while all remaining agent components and interaction protocols are shared. On the physics-only HiPhO benchmark~\citep{2025hipho}, the system employs the Physics-Verifier in a setting consistent with prior physics-focused evaluations. On the multidisciplinary FrontierScience-Olympiad benchmark~\citep{frontierscience}, the same pipeline dynamically routes problems to the appropriate domain verifier and solver configuration. These experiments demonstrate that combining visual grounding with domain-adaptive agentic reasoning supports effective performance across physics, chemistry, and biology Olympiad-level problems.

\section{Experiment}

% \subsection{Settings and Baselines}
%%%%%%%%%%%%%%%%%%%%%%%%%%%%%%%%%%%%%%
\subsection{Experimental Setup}

\textbf{Test Dataset.} 
To rigorously evaluate performance on advanced physical reasoning tasks, we introduce the HiPhO benchmark~\citep{2025hipho}, the first dataset dedicated exclusively to High School Physics Olympiads. HiPhO aggregates 13 major exams administered between 2024 and 2025, spanning seven prestigious international and regional competition series: IPhO, APhO, EuPhO, NBPhO, PanPhO, PanMechanics, and F=MA. The inclusion criteria prioritized both global influence and the availability of granular human performance data to facilitate direct human-AI comparison\footnote{Notable competitions such as CPhO, USAPhO, and APhO-2024 were excluded due to the unavailability of complete official contestant score distributions.}.

\textbf{Comparison Models.}
We compared against 39 representative models, including 13 closed-source and 26 open-source models, selected to reflect the current frontier of LLMs and VLMs :  
\vspace{-1mm}
\begin{itemize}[left=0pt]
    \item \textbf{Closed-source models:} Gemini-3-Pro \citep{gemini-3-pro}, GPT-5.2 \citep{GPT-5.2}, GPT-5 \citep{GPT-5}, o3 \citep{o3_o4-mini}, o4-mini (high) \citep{o3_o4-mini}, o4-mini \citep{o3_o4-mini}, GPT-4o \citep{GPT-4o}, Gemini-2.5-Pro \citep{Gemini-2.5}, Gemini-2.5-Flash-Thinking \citep{Gemini-2.5}, Grok-4 \citep{Grok-4}, Claude-4-Sonnet-Thinking \citep{Claude-3.7-Sonnet}, Claude-4-Sonnet \citep{Claude-3.7-Sonnet}, Mistral-Medium-3 \citep{Mistral-Medium-3}.
    \vspace{1mm}
    \item \textbf{Open-source models:} DeepSeek-V3.2-Thinking \citep{DeepSeekAI2025DeepSeekV32PT}, Qwen3-VL Series \citep{bai2025qwen3vltechnicalreport}, Intern-S1 \citep{Intern-S1}, InternVL3 Series \citep{InternVL3}, Qwen2.5-VL Series \citep{Qwen2.5-VL}, GLM-4.5V \citep{GLM-4.5V}, DeepSeek-VL2 \citep{DeepSeek-VL2}, LLaMA4-Scout-17B \citep{LLaMA4-Scout}, Phi-4-multimodal \citep{Phi-4-multimodal}, P1 Series \citep{chen2025p1masteringphysicsolympiads}, GPT-OSS-120B \citep{2025GPT-OSS}, Kimi-K2-Thinking \citep{2025Kimi-K2}, Kimi-K2-Instruct \citep{2025Kimi-K2}, DeepSeek-R1 \citep{guo2025deepseek-r1}, DeepSeek-V3 \citep{2024DeepSeek-V3}, Qwen3 Series \citep{yang2025qwen3}.  
\end{itemize}
\vspace{-1mm}

\textbf{Model Configurations.}
All models were evaluated under a standardized inference setup following the HiPhO protocol. The temperature was fixed at $0.6$, and the maximum token limit was set as large as permitted by each model. For every problem, we conducted $8$ independent inference runs and computed the average score per problem. These averages were then aggregated across problems to yield the final exam score for each Olympiad.

\textbf{Evaluation Method.} Adhering to the HiPhO benchmark protocol~\citep{2025hipho}, we employ \texttt{Gemini-2.5-Flash} as the automated evaluator. The evaluation framework mirrors official Olympiad grading by integrating both \emph{answer-level} verification and \emph{step-level} reasoning analysis. For each problem, the final score is determined as the maximum of the answer-based and step-based evaluations. This mechanism ensures consistency with human adjudication: a correct final answer gains full credit, while rigorous intermediate derivations earn partial points even if the final result is incorrect. Consequently, we report the cumulative \textit{exam score} rather than standard accuracy, facilitating a direct comparison between model performance and official human medal thresholds.

%%%%%%%%%%%%%%%%%%%%%%%%%%%%%%%%%%%%%%

\subsection{Evaluation on Physics Olympiads}

\textbf{Superior Single-Model Performance.} Evaluation results are presented in Table~\ref{tab:hipho_all}, verifying the effectiveness of our RL training strategy. As noted above, the P1-VL model series is trained purely through reinforcement learning. 
\vspace{-2mm}
\begin{itemize}[left=0pt]
    \item \textbf{P1-VL-235B-A22B} ranks \textbf{3rd} among all evaluated models, following only Gemini-3-Pro and GPT-5.2. With 12 gold and 1 silver medals (average score: 39.3), it surpasses the closed-source models such as Gemini-2.5-Pro (37.7), GPT-5 (37.4), and Grok-4 (35.8). Notably, the standalone P1-VL outperforms the agent-augmented baseline (\texttt{P1-235B-A22B} + \texttt{PhysicsMinions}, 38.4), proving that intrinsic visual reasoning capabilities can surpass purely agentic workflows in complex physical scenarios, which incorporates image caption information. It shows that our models could effectively perceive the image information and bridge it to the scientific reasoning (case study for Text+Variable Figure is shown in Appendix \ref{sec:case1}). 
    Compared with its base model (Qwen3-VL-235B-A22B-Thinking, 33.9), it improves 5.4 points on average. 
    \vspace{2mm}
    \item \textbf{P1-VL-30B-A3B} achieves 9 gold and 4 silver medals with an average score of 35.0. It ranks third among open-source models, behind only DeepSeek-V3.2-Thinking and P1-235B-A22B. Despite its smaller size, it outperforms larger baselines such as  Qwen3-VL-235B-A22B-Thinking (33.9) and Qwen3-235B-A22B-Thinking-2507 (33.5), demonstrating significant parameter efficiency. It also outperforms some closed-source models such as o4-mini and Gemini-2.5-Flash-Thinking, showing excellent performance on scientific reasoning. 
    \vspace{-2mm}
\end{itemize}

\textbf{Agentic Augmentation.} 
With the agentic augmentation of \textit{PhysicsMinions}, the average score of \texttt{P1-VL-235B-A22B} improves from 39.3 to 40.9, propelling it to the \textbf{2nd rank globally} and surpassing the closed-source GPT-5.2 (40.6). This combined system establishes new state-of-the-art performance on three physics Olympiads, including PanPhO 2025 (66.5 vs 66.3), PanPhO 2024 (83.3 vs 82.5), and PanMechanics 2024 (84.8 vs 82.3). These gains underscore the efficacy of the "model + system" paradigm, demonstrating how multi-agent collaboration significantly amplifies complex scientific reasoning capabilities.

\begin{table}[th]
\centering
\vspace{-5pt}
\caption{Evaluation results on the HiPhO benchmark (13 physics Olympiads from 2024--2025) using the \textit{exam score} metric. \medalbox{Gold!50}{Gold}, \medalbox{Silver!70}{Silver} and \medalbox{Bronze!40}{Bronze} indicate scores above the respective thresholds. Models are ranked by average scores; \textbf{bold} is the highest score, and \underline{underline} is the second highest. Here, only the theoretical parts of exams are used, hence Full Mark (Model) $\leq$ Full Mark (Human).}
\vspace{-4pt}
\label{tab:hipho_all}
\small
\resizebox{\textwidth}{!}{%
\setlength{\tabcolsep}{2.3pt}

\begin{tabular}{lccccccccccccc|c|ccc}
\toprule
\textbf{Physics Olympiad} & \multicolumn{2}{c}{\textbf{IPhO}} & \textbf{APhO} & \multicolumn{2}{c}{\textbf{EuPhO}} & \multicolumn{2}{c}{\textbf{NBPhO}} & \multicolumn{6}{c|}{\textbf{PanPhO}~~
\textbf{\footnotesize PanMechanics} ~~ \textbf{F=MA}} & \textbf{Avg.} & \multicolumn{3}{c}{\textbf{Medal}} \\
\textbf{Year} &2025&2024&2025&2025&2024&2025&2024&2025&2024&2025&2024&2025&2024& & \multicolumn{3}{c}{\textbf{Table}} \\ \midrule
Full Mark (Human) & 30.0 & 30.0 & 30.0 & 30.0 & 30.0 & 72.0 & 72.0 & 100.0 & 100.0 & 100.0 & 100.0 & 25.0 & 25.0 & 57.2 \\
Full Mark (Model) & 29.4 & 29.3 & 30.0 & 29.0 & 28.0 & 43.5 & 50.0 & 100.0 & 98.0 & 100.0 & 100.0 & 25.0 & 25.0 & 52.9 \\
Top-1 Score (Human) & 29.2 & 29.4 & 30.0 & 27.0 & 30.0 & 53.2 & 40.8 & 81.0 & 66.5 & 62.0 & 51.0 & 25.0 & 24.0 & 42.2 \\
Top-1 Score (Compared Models) & 25.2 & 25.9 & 28.4 & 21.0 & 23.9 & 37.1 & 43.8 & 66.3 & 82.5 & 86.0 & 82.3 & 23.8 & 23.9 & 43.9 \\
\rowcolor{Gold!40}
Gold Medal   & 19.7 & 20.8 & 23.3 & 16.5 & 20.4 & 28.6 & 26.5 & 41.5 & 52.0 & 52.0 & 51.0 & 15.0 & 14.0 & \cellcolor{white}29.3 & \multicolumn{3}{>{\cellcolor{white}}l}{\textcolor{Gold}{\faMedal}}  \\
\rowcolor{Silver!60}
Silver Medal & 12.1 & 11.1 & 18.7 & 9.8 & 14.2 & 20.1 & 19.4 & 28.5 & 37.5 & 36.0 & 26.0 & 11.0 & 12.0 & \cellcolor{white}19.7 & \multicolumn{3}{>{\cellcolor{white}}c}{\textcolor{Silver}{\faMedal}}  \\
\rowcolor{Bronze!40}
Bronze Medal & 7.2  & 3.6  & 13.1 & 5.8  & 8.9  & 15.2  & 13.5  & 14.5 & 16.0 & 20.0 & 12.0 & 9.0 & 10.0 & \cellcolor{white}11.4 & \multicolumn{3}{>{\cellcolor{white}}r}{\textcolor{Bronze}{\faMedal}}  \\ \midrule
Gemini-3-Pro (high) &\cellcolor{Gold!35}\textbf{25.2}&\cellcolor{Gold!35}{25.2}&\cellcolor{Gold!35}\underline{28.3}&\cellcolor{Gold!35}\textbf{21.0}&\cellcolor{Gold!35}\textbf{23.9}&\cellcolor{Gold!35}\textbf{37.1}&\cellcolor{Gold!35}\textbf{43.8}&\cellcolor{Gold!35}\underline{66.3}&\cellcolor{Gold!35}{76.2}&\cellcolor{Gold!35}\textbf{86.0}&\cellcolor{Gold!35}74.8&\cellcolor{Gold!35}\textbf{23.8}&\cellcolor{Gold!35}\textbf{23.9}&\textbf{42.7}& ~\mc{Gold}{13} & \mc{Silver}{0} & \mc{Bronze}{0}\\
\textcolor[HTML]{78206E}{\textbf{P1-VL-235B-A22B+PhysicsMinions}} &\cellcolor{Gold!35}22.1&\cellcolor{Gold!35}\underline{25.8}&\cellcolor{Gold!35}27.1&\cellcolor{Silver!60}13.7&\cellcolor{Gold!35}22.9&\cellcolor{Gold!35}31.8&\cellcolor{Gold!35}32.6&\cellcolor{Gold!35}\textbf{66.5}&\cellcolor{Gold!35}\textbf{83.3}&\cellcolor{Gold!35}{76.0}&\cellcolor{Gold!35}\textbf{84.8}&\cellcolor{Gold!35}23.1&\cellcolor{Gold!35}21.8&\underline{40.9}& ~\mc{Gold}{12} & \mc{Silver}{1} & \mc{Bronze}{0}\\
GPT-5.2 (high) &\cellcolor{Gold!35}22.3&\cellcolor{Gold!35}25.5&\cellcolor{Gold!35}\textbf{28.4}&\cellcolor{Silver!60}\underline{15.7}&\cellcolor{Gold!35}\textbf{23.9}&\cellcolor{Gold!35}\underline{35.8}&\cellcolor{Gold!35}\underline{37.1}&\cellcolor{Gold!35}{59.8}&\cellcolor{Gold!35}\underline{82.5}&\cellcolor{Gold!35}67.6&\cellcolor{Gold!35}\underline{82.3}&\cellcolor{Gold!35}\underline{23.4}&\cellcolor{Gold!35}\underline{23.5}&{40.6}& ~\mc{Gold}{12} & \mc{Silver}{1} & \mc{Bronze}{0}\\
\textcolor[HTML]{78206E}{\textbf{P1-VL-235B-A22B }}&\cellcolor{Gold!35}21.0&\cellcolor{Gold!35}25.3&\cellcolor{Gold!35}26.9&\cellcolor{Silver!60}10.7&\cellcolor{Gold!35}21.5&\cellcolor{Gold!35}32.5&\cellcolor{Gold!35}31.9&\cellcolor{Gold!35}61.7&\cellcolor{Gold!35}{79.8}&\cellcolor{Gold!35}{73.3}&\cellcolor{Gold!35}{82.0}&\cellcolor{Gold!35}22.5&\cellcolor{Gold!35}21.9&39.3& ~\mc{Gold}{12} & \mc{Silver}{1} & \mc{Bronze}{0}\\
\textcolor[HTML]{6c25be}{\textbf{P1-235B-A22B+PhysicsMinions}} &\cellcolor{Gold!35}\underline{23.2}&\cellcolor{Gold!35}25.2&\cellcolor{Gold!35}28.0&\cellcolor{Silver!60}12.4&\cellcolor{Gold!35}\underline{23.5}&\cellcolor{Gold!35}31.9&\cellcolor{Gold!35}35.4&\cellcolor{Gold!35}57.7&\cellcolor{Gold!35}67.0&\cellcolor{Gold!35}\underline{77.5}&\cellcolor{Gold!35}74.8&\cellcolor{Gold!35}21.5&\cellcolor{Gold!35}20.5&38.4& ~\mc{Gold}{12} & \mc{Silver}{1} & \mc{Bronze}{0}\\
Gemini-2.5-Pro &\cellcolor{Gold!35}22.7&\cellcolor{Gold!35}\textbf{25.9}&\cellcolor{Gold!35}27.9&\cellcolor{Silver!60}14.9&\cellcolor{Gold!35}21.8&\cellcolor{Gold!35}32.3&\cellcolor{Gold!35}35.9&\cellcolor{Gold!35}60.3&\cellcolor{Gold!35}64.1&\cellcolor{Gold!35}69.5&\cellcolor{Gold!35}70.2&\cellcolor{Gold!35}22.8&\cellcolor{Gold!35}22.0&37.7& ~\mc{Gold}{12} & \mc{Silver}{1} & \mc{Bronze}{0}\\
GPT-5 &\cellcolor{Gold!35}22.3&\cellcolor{Silver!60}20.2&\cellcolor{Gold!35}27.0&\cellcolor{Silver!60}10.3&\cellcolor{Gold!35}21.7&\cellcolor{Gold!35}32.9&\cellcolor{Gold!35}32.8&\cellcolor{Gold!35}55.9&\cellcolor{Gold!35}69.8&\cellcolor{Gold!35}69.4&\cellcolor{Gold!35}79.0&\cellcolor{Gold!35}22.4&\cellcolor{Gold!35}22.4&37.4& ~\mc{Gold}{11} & \mc{Silver}{2} & \mc{Bronze}{0} \\
DeepSeek-V3.2-Thinking		&	\cellcolor{Gold!35}20.2 &	\cellcolor{Gold!35}24.9 &\cellcolor{Gold!35}26.1 &	\cellcolor{Bronze!40}6.7 &	\cellcolor{Silver!60}18.9 &	\cellcolor{Gold!35}33.3& 	\cellcolor{Gold!35}29.7 &	\cellcolor{Gold!35}57.3 	&\cellcolor{Gold!35}64.9 	&\cellcolor{Gold!35}78.6 &	\cellcolor{Gold!35}78.0 &	\cellcolor{Gold!35}21.5 	&\cellcolor{Gold!35}19.9 &	36.9&	~\mc{Gold}{11}&	\mc{Silver}1&	\mc{Bronze}1 \\
\textcolor[HTML]{6c25be}{\textbf{P1-235B-A22B}} &\cellcolor{Gold!35}21.2&\cellcolor{Gold!35}24.7&\cellcolor{Gold!35}27.4&\cellcolor{Silver!60}10.8&\cellcolor{Gold!35}23.0&\cellcolor{Gold!35}31.8&\cellcolor{Gold!35}28.4&\cellcolor{Gold!35}54.7&\cellcolor{Gold!35}56.7&\cellcolor{Gold!35}74.7&\cellcolor{Gold!35}72.9&\cellcolor{Gold!35}20.9&\cellcolor{Gold!35}19.4&35.9& ~\mc{Gold}{12} & \mc{Silver}{1} & \mc{Bronze}{0}\\
Grok-4 &\cellcolor{Silver!60}18.7&\cellcolor{Gold!35}23.5&\cellcolor{Gold!35}25.0&\cellcolor{Silver!60}11.5&\cellcolor{Gold!35}20.5&\cellcolor{Silver!60}25.8&\cellcolor{Gold!35}29.3&\cellcolor{Gold!35}45.0&\cellcolor{Gold!35}75.4&\cellcolor{Gold!35}72.1&\cellcolor{Gold!35}78.6&\cellcolor{Gold!35}19.8&\cellcolor{Gold!35}19.8&35.8& ~\mc{Gold}{10} & \mc{Silver}{3} & \mc{Bronze}{0}\\
o3 &\cellcolor{Silver!60}15.7&\cellcolor{Gold!35}23.7&\cellcolor{Gold!35}25.9&\cellcolor{Silver!60}11.4&\cellcolor{Gold!35}21.6&\cellcolor{Gold!35}34.1&\cellcolor{Gold!35}33.5&\cellcolor{Gold!35}47.3&\cellcolor{Gold!35}55.9&\cellcolor{Gold!35}71.4&\cellcolor{Gold!35}75.6&\cellcolor{Gold!35}22.0&\cellcolor{Gold!35}20.6&35.3& ~\mc{Gold}{11} & \mc{Silver}{2} & \mc{Bronze}{0}\\
\textcolor[HTML]{78206E}{\textbf{P1-VL-30B-A3B}} &\cellcolor{Silver!60}17.9&\cellcolor{Gold!35}22.5&\cellcolor{Gold!35}24.6&\cellcolor{Silver!60}10.6&\cellcolor{Silver!60}19.0&\cellcolor{Gold!35}30.2&\cellcolor{Silver!60}24.6&\cellcolor{Gold!35}50.8&\cellcolor{Gold!35}58.6&\cellcolor{Gold!35}79.1&\cellcolor{Gold!35}75.5&\cellcolor{Gold!35}20.8&\cellcolor{Gold!35}20.3&35.0& ~\mc{Gold}{9} & \mc{Silver}{4} & \mc{Bronze}{0}\\
% \textcolor[HTML]{6c25be}{\textbf{P1-30B-A3B-0930}} 
%  &	18.5 &	23.6 	&24.7 	&8.4 	&19.0 &	31.0 &	26.4 	&57.0 	&50.1 &	75.5 	&76.9 &	20.3 &	19.9 	&34.7 	& ~\mc{Gold}{8} & \mc{Silver}{4} & \mc{Bronze}{1} \\
Qwen3-VL-235B-A22B-Thinking &\cellcolor{Gold!35}19.8&\cellcolor{Gold!35}23.4&\cellcolor{Gold!35}25.4&\cellcolor{Silver!60}10.8&\cellcolor{Silver!60}19.0&\cellcolor{Gold!35}29.2&\cellcolor{Silver!60}25.9&\cellcolor{Gold!35}48.3&\cellcolor{Gold!35}54.0&\cellcolor{Gold!35}64.3&\cellcolor{Gold!35}79.9&\cellcolor{Gold!35}21.4&\cellcolor{Gold!35}19.9&33.9& ~\mc{Gold}{10} & \mc{Silver}{3} & \mc{Bronze}{0}\\
Qwen3-235B-A22B-Thinking-2507 &\cellcolor{Silver!60}17.1&\cellcolor{Gold!35}23.0&\cellcolor{Gold!35}26.2&\cellcolor{Silver!60}10.9&\cellcolor{Gold!35}20.4&\cellcolor{Gold!35}33.6&\cellcolor{Gold!35}28.1&\cellcolor{Gold!35}44.7&\cellcolor{Silver!60}51.8&\cellcolor{Gold!35}69.1&\cellcolor{Gold!35}72.9&\cellcolor{Gold!35}18.5&\cellcolor{Gold!35}18.9&33.5& ~\mc{Gold}{10} & \mc{Silver}{3} & \mc{Bronze}{0}\\
\textcolor[HTML]{6c25be}{\textbf{P1-30B-A3B}} &\cellcolor{Silver!60}18.5&\cellcolor{Gold!35}22.3&\cellcolor{Gold!35}25.4&\cellcolor{Bronze!40}7.4&\cellcolor{Silver!60}18.8&\cellcolor{Gold!35}29.2&\cellcolor{Silver!60}24.0&\cellcolor{Gold!35}47.0&\cellcolor{Silver!60}51.4&\cellcolor{Gold!35}69.5&\cellcolor{Gold!35}69.1&\cellcolor{Gold!35}19.6&\cellcolor{Gold!35}20.0&32.5& ~\mc{Gold}{8} & \mc{Silver}{4} & \mc{Bronze}{1}\\

GPT-OSS-120B (high)		&		\cellcolor{Gold!35}20.5 	&\cellcolor{Gold!35}22.1 	&\cellcolor{Gold!35}25.8 	&\cellcolor{Silver!60}12.8 	&\cellcolor{Gold!35}23.0 	&\cellcolor{Gold!35}32.4 	&\cellcolor{Gold!35}32.4 	&\cellcolor{Gold!35}48.8 	&\cellcolor{Gold!35}52.4 	&\cellcolor{Gold!35}54.7 	&\cellcolor{Gold!35}56.1 	&\cellcolor{Gold!35}20.5 	&\cellcolor{Gold!35}19.2 	&32.4 	&~\mc{Gold}{12}	&\mc{Silver}1	&\mc{Bronze}0 \\

o4-mini (high) &\cellcolor{Silver!60}16.0&\cellcolor{Gold!35}23.7&\cellcolor{Silver!60}22.9&\cellcolor{Silver!60}12.0&\cellcolor{Silver!60}20.1&\cellcolor{Silver!60}27.4&\cellcolor{Gold!35}29.8&\cellcolor{Silver!60}41.4&\cellcolor{Silver!60}50.9&\cellcolor{Gold!35}69.1&\cellcolor{Gold!35}67.3&\cellcolor{Gold!35}18.6&\cellcolor{Gold!35}18.8&32.2& ~\mc{Gold}{6} & \mc{Silver}{7} & \mc{Bronze}{0}\\
Gemini-2.5-Flash-Thinking &\cellcolor{Gold!35}20.2&\cellcolor{Gold!35}23.9&\cellcolor{Gold!35}27.4&\cellcolor{Silver!60}13.2&\cellcolor{Gold!35}21.9&\cellcolor{Gold!35}29.0&\cellcolor{Gold!35}29.3&\cellcolor{Gold!35}44.6&\cellcolor{Gold!35}54.9&\cellcolor{Gold!35}60.5&\cellcolor{Gold!35}55.9&\cellcolor{Gold!35}17.8&\cellcolor{Gold!35}19.1&32.1& ~\mc{Gold}{12} & \mc{Silver}{1} & \mc{Bronze}{0}\\
DeepSeek-R1 &\cellcolor{Silver!60}18.5&\cellcolor{Gold!35}24.6&\cellcolor{Gold!35}25.4&\cellcolor{Silver!60}10.8&\cellcolor{Gold!35}21.4&\cellcolor{Silver!60}26.3&\cellcolor{Silver!60}20.5&\cellcolor{Gold!35}42.2&\cellcolor{Silver!60}47.4&\cellcolor{Gold!35}65.4&\cellcolor{Gold!35}72.5&\cellcolor{Gold!35}18.3&\cellcolor{Gold!35}18.5&31.7& ~\mc{Gold}{8} & \mc{Silver}{5} & \mc{Bronze}{0} \\

o4-mini &\cellcolor{Silver!60}15.4&\cellcolor{Gold!35}22.9&\cellcolor{Silver!60}22.8&\cellcolor{Silver!60}10.1&\cellcolor{Gold!35}20.9&\cellcolor{Silver!60}26.9&\cellcolor{Gold!35}27.3&\cellcolor{Silver!60}39.4&\cellcolor{Silver!60}47.1&\cellcolor{Gold!35}64.2&\cellcolor{Gold!35}62.5&\cellcolor{Gold!35}18.6&\cellcolor{Gold!35}18.5&30.5& ~\mc{Gold}{7} & \mc{Silver}{6} & \mc{Bronze}{0} \\

% Qwen3-235B-A22B &\cellcolor{Silver!60}17.8&\cellcolor{Gold!35}23.8&\cellcolor{Gold!35}26.0&\cellcolor{Bronze!40}9.2&\cellcolor{Gold!35}21.5&\cellcolor{Silver!60}28.4&\cellcolor{Gold!35}31.1&\cellcolor{Gold!35}42.3&\cellcolor{Silver!60}49.1&\cellcolor{Gold!35}63.1&\cellcolor{Silver!60}44.6&\cellcolor{Gold!35}18.4&\cellcolor{Gold!35}17.4&30.2& ~\mc{Gold}{8} & \mc{Silver}{4} & \mc{Bronze}{1} \\
%
Kimi-K2-Thinking* &\cellcolor{Silver!60}19.1&\cellcolor{Gold!35}22.8&\cellcolor{Silver!60}22.4&5.2&\cellcolor{Gold!35}20.9&\cellcolor{Silver!60}26.0&\cellcolor{Silver!60}20.6&\cellcolor{Gold!35}52.5&\cellcolor{Silver!60}51.3&\cellcolor{Silver!60}47.3&\cellcolor{Gold!35}64.8&\cellcolor{Gold!35}20.6&\cellcolor{Gold!35}19.7&30.2& ~\mc{Gold}{6} & \mc{Silver}{6} & \mc{Bronze}{0} \\
Claude-4-Sonnet-Thinking &\cellcolor{Silver!60}19.0&\cellcolor{Gold!35}22.0&\cellcolor{Gold!35}24.8&\cellcolor{Bronze!40}9.7&\cellcolor{Gold!35}20.5&\cellcolor{Silver!60}28.1&\cellcolor{Silver!60}25.6&\cellcolor{Gold!35}43.1&\cellcolor{Silver!60}39.3&\cellcolor{Gold!35}57.4&\cellcolor{Gold!35}61.8&\cellcolor{Gold!35}19.2&\cellcolor{Gold!35}20.1&30.0& ~\mc{Gold}{8} & \mc{Silver}{4} & \mc{Bronze}{1}\\
Qwen3-30B-A3B-Thinking-2507 &	\cellcolor{Silver!60}15.6 &	\cellcolor{Silver!60}19.7 &	\cellcolor{Gold!35}23.5 &	\cellcolor{Bronze!40}7.4 &\cellcolor{Silver!60}16.4 	&\cellcolor{Gold!35}28.6 	&\cellcolor{Silver!60}22.2 &	\cellcolor{Silver!60}40.5 &	\cellcolor{Silver!60}43.8 &	\cellcolor{Gold!35}67.7 	&\cellcolor{Gold!35}66.5 	&\cellcolor{Gold!35}18.3 &	\cellcolor{Gold!35}18.0 	&29.9 &~\mc{Gold}{6} & \mc{Silver}{6} & \mc{Bronze}{1}\\

Qwen3-VL-30B-A3B-Thinking &\cellcolor{Silver!60}15.7&\cellcolor{Silver!60}19.3&\cellcolor{Gold!35}25.3&\cellcolor{Silver!60}11.7&\cellcolor{Silver!60}20.2&\cellcolor{Gold!35}29.7&\cellcolor{Gold!35}28.3&\cellcolor{Gold!35}41.6&\cellcolor{Silver!60}45.9&\cellcolor{Gold!35}57.6&\cellcolor{Gold!35}51.4&\cellcolor{Gold!35}18.9&\cellcolor{Gold!35}20.7&29.7& ~\mc{Gold}{8} & \mc{Silver}{5} & \mc{Bronze}{0}\\
Qwen3-32B &\cellcolor{Silver!60}15.7&\cellcolor{Silver!60}19.3&\cellcolor{Gold!35}23.9&\cellcolor{Silver!60}9.8&\cellcolor{Gold!35}21.2&\cellcolor{Gold!35}28.9&\cellcolor{Silver!60}24.1&\cellcolor{Silver!60}36.6&\cellcolor{Silver!60}41.8&\cellcolor{Gold!35}67.0&\cellcolor{Gold!35}59.2&\cellcolor{Gold!35}18.9&\cellcolor{Gold!35}16.6&29.5& ~\mc{Gold}{7} & \mc{Silver}{6} & \mc{Bronze}{0} \\
Kimi-K2-Instruct &\cellcolor{Silver!60}16.5&\cellcolor{Silver!60}19.8&\cellcolor{Gold!35}24.2&\cellcolor{Silver!60}11.0&\cellcolor{Silver!60}16.9&\cellcolor{Silver!60}26.5&\cellcolor{Silver!60}26.2&\cellcolor{Silver!60}35.9&\cellcolor{Silver!60}41.8&\cellcolor{Gold!35}65.9&\cellcolor{Gold!35}58.9&\cellcolor{Gold!35}16.0&\cellcolor{Gold!35}18.2&29.1& ~\mc{Gold}{5} & \mc{Silver}{8} & \mc{Bronze}{0} \\
GPT-OSS-120B &\cellcolor{Silver!60}16.9&\cellcolor{Gold!35}21.4&\cellcolor{Silver!60}22.8&\cellcolor{Bronze!40}9.1&\cellcolor{Silver!60}19.9&\cellcolor{Silver!60}26.0&\cellcolor{Silver!60}25.8&\cellcolor{Silver!60}37.4&\cellcolor{Silver!60}41.8&\cellcolor{Gold!35}57.1&\cellcolor{Gold!35}59.7&\cellcolor{Gold!35}17.8&\cellcolor{Gold!35}17.6&28.7& ~\mc{Gold}{5} & \mc{Silver}{7} & \mc{Bronze}{1} \\
Intern-S1 &\cellcolor{Silver!60}15.9&\cellcolor{Silver!60}14.2&\cellcolor{Silver!60}21.7&\cellcolor{Bronze!40}9.0&\cellcolor{Silver!60}16.6&\cellcolor{Silver!60}23.0&\cellcolor{Silver!60}20.5&\cellcolor{Silver!60}41.1&\cellcolor{Silver!60}50.3&\cellcolor{Gold!35}60.4&\cellcolor{Gold!35}57.4&\cellcolor{Gold!35}18.4&\cellcolor{Gold!35}19.5&28.3& ~\mc{Gold}{4} & \mc{Silver}{8} & \mc{Bronze}{1} \\
Claude-4-Sonnet &\cellcolor{Silver!60}15.7&\cellcolor{Silver!60}19.2&\cellcolor{Silver!60}22.8&\cellcolor{Bronze!40}9.5&\cellcolor{Silver!60}16.5&\cellcolor{Silver!60}27.5&\cellcolor{Silver!60}21.3&\cellcolor{Silver!60}40.4&\cellcolor{Silver!60}43.3&\cellcolor{Silver!60}46.5&\cellcolor{Silver!60}48.5&\cellcolor{Gold!35}16.8&\cellcolor{Gold!35}16.5&26.5& ~\mc{Gold}{2} & \mc{Silver}{10} & \mc{Bronze}{1}\\
%
% Qwen3-30B-A3B &\cellcolor{Silver!60}13.6&\cellcolor{Silver!60}15.4&\cellcolor{Silver!60}22.7&\cellcolor{Silver!60}9.8&\cellcolor{Silver!60}16.5&\cellcolor{Silver!60}24.7&\cellcolor{Silver!60}21.7&\cellcolor{Silver!60}31.9&\cellcolor{Silver!60}39.5&\cellcolor{Silver!60}49.9&\cellcolor{Silver!60}45.0&\cellcolor{Gold!35}15.5&\cellcolor{Gold!35}15.0&24.7& ~\mc{Gold}{2} & \mc{Silver}{11} & \mc{Bronze}{0} \\
%
DeepSeek-V3 &\cellcolor{Silver!60}13.6&\cellcolor{Silver!60}16.4&\cellcolor{Silver!60}22.1&\cellcolor{Bronze!40}7.1&\cellcolor{Silver!60}17.2&\cellcolor{Silver!60}21.1&\cellcolor{Bronze!40}17.3&\cellcolor{Silver!60}37.2&\cellcolor{Bronze!40}35.0&\cellcolor{Silver!60}48.4&\cellcolor{Silver!60}46.5&\cellcolor{Silver!60}14.1&\cellcolor{Gold!35}15.6&24.0& ~\mc{Gold}{1} & \mc{Silver}{9} & \mc{Bronze}{3} \\
Mistral-Medium-3 &\cellcolor{Silver!60}14.2&\cellcolor{Silver!60}14.1&\cellcolor{Silver!60}19.9&\cellcolor{Bronze!40}8.5&\cellcolor{Bronze!40}12.2&\cellcolor{Silver!60}20.4&\cellcolor{Silver!60}19.6&\cellcolor{Silver!60}30.8&\cellcolor{Bronze!40}28.6&\cellcolor{Bronze!40}32.9&\cellcolor{Silver!60}36.1&\cellcolor{Silver!60}13.9&\cellcolor{Gold!35}14.1&20.4& ~\mc{Gold}{1} & \mc{Silver}{8} & \mc{Bronze}{4}\\
InternVL3-78B-Instruct &\cellcolor{Silver!60}12.9&\cellcolor{Silver!60}12.5&\cellcolor{Bronze!40}17.7&\cellcolor{Bronze!40}7.5&\cellcolor{Silver!60}15.2&\cellcolor{Silver!60}22.3&\cellcolor{Silver!60}22.5&\cellcolor{Bronze!40}26.2&\cellcolor{Bronze!40}27.4&\cellcolor{Bronze!40}21.1&\cellcolor{Silver!60}27.1&\cellcolor{Silver!60}12.0&\cellcolor{Silver!60}13.0&18.3& ~\mc{Gold}{0} & \mc{Silver}{8} & \mc{Bronze}{5}\\
GLM-4.5V &\cellcolor{Bronze!40}11.9&\cellcolor{Bronze!40}4.4&\cellcolor{Bronze!40}16.2&\cellcolor{Bronze!40}8.7&\cellcolor{Bronze!40}14.1&\cellcolor{Bronze!40}19.5&\cellcolor{Bronze!40}14.0&\cellcolor{Bronze!40}18.5&\cellcolor{Bronze!40}16.0&\cellcolor{Silver!60}47.8&\cellcolor{Silver!60}39.0&\cellcolor{Silver!60}13.0&\cellcolor{Silver!60}13.8&18.2& ~\mc{Gold}{0} & \mc{Silver}{4} & \mc{Bronze}{9} \\
LLaMA4-Scout-17B &\cellcolor{Bronze!40}9.7&\cellcolor{Bronze!40}9.5&\cellcolor{Bronze!40}13.1&5.4&\cellcolor{Bronze!40}10.4&\cellcolor{Silver!60}22.8&12.8&\cellcolor{Bronze!40}26.6&\cellcolor{Bronze!40}24.1&\cellcolor{Bronze!40}35.4&\cellcolor{Silver!60}34.5&8.1&6.4&16.8& ~\mc{Gold}{0} & \mc{Silver}{2} & \mc{Bronze}{7}\\
GPT-4o &\cellcolor{Bronze!40}10.2&\cellcolor{Bronze!40}9.4&\cellcolor{Bronze!40}15.1&\cellcolor{Bronze!40}6.8&\cellcolor{Bronze!40}9.2&\cellcolor{Bronze!40}16.4&11.7&\cellcolor{Bronze!40}27.8&\cellcolor{Bronze!40}22.8&\cellcolor{Bronze!40}28.2&\cellcolor{Silver!60}26.5&\cellcolor{Gold!35}15.0&\cellcolor{Bronze!40}10.9&16.2& ~\mc{Gold}{1} & \mc{Silver}{1} & \mc{Bronze}{10}\\
Qwen3-8B &\cellcolor{Bronze!40}10.6&\cellcolor{Silver!60}12.7&11.5&\cellcolor{Bronze!40}7.1&\cellcolor{Bronze!40}11.9&\cellcolor{Silver!60}20.1&\cellcolor{Bronze!40}17.3&\cellcolor{Bronze!40}26.3&\cellcolor{Bronze!40}22.3&\cellcolor{Bronze!40}21.8&\cellcolor{Bronze!40}22.8&\cellcolor{Bronze!40}10.8&\cellcolor{Bronze!40}10.0&15.8& ~\mc{Gold}{0} & \mc{Silver}{2} & \mc{Bronze}{10} \\
Qwen2.5-VL-32B-Instruct &\cellcolor{Bronze!40}9.9&\cellcolor{Bronze!40}8.2&\cellcolor{Bronze!40}16.5&\cellcolor{Bronze!40}6.9&\cellcolor{Bronze!40}10.0&\cellcolor{Bronze!40}15.3&\cellcolor{Bronze!40}14.4&\cellcolor{Bronze!40}22.5&\cellcolor{Bronze!40}22.4&\cellcolor{Bronze!40}28.1&\cellcolor{Silver!60}29.9&7.6&4.6&15.1& ~\mc{Gold}{0} & \mc{Silver}{1} & \mc{Bronze}{10} \\
Qwen2.5-VL-72B-Instruct &\cellcolor{Bronze!40}10.6&\cellcolor{Bronze!40}7.2&\cellcolor{Bronze!40}13.6&\cellcolor{Bronze!40}6.1&8.1&13.3&11.4&\cellcolor{Bronze!40}26.8&\cellcolor{Bronze!40}18.2&\cellcolor{Bronze!40}24.0&\cellcolor{Silver!60}28.5&\cellcolor{Silver!60}13.5&9.8&14.7& ~\mc{Gold}{0} & \mc{Silver}{2} & \mc{Bronze}{7} \\
InternVL3-38B-Instruct &\cellcolor{Bronze!40}8.9&\cellcolor{Bronze!40}7.8&12.3&\cellcolor{Bronze!40}6.1&8.3&14.0&10.6&\cellcolor{Bronze!40}24.1&\cellcolor{Bronze!40}20.4&\cellcolor{Bronze!40}27.5&\cellcolor{Bronze!40}24.8&8.2&6.8&13.8& ~\mc{Gold}{0} & \mc{Silver}{0} & \mc{Bronze}{7}\\
InternVL3-9B-Instruct &4.7&\cellcolor{Bronze!40}3.7&7.2&4.2&4.1&9.4&6.0&12.4&11.0&11.6&\cellcolor{Bronze!40}16.3&6.4&6.2&7.9& ~\mc{Gold}{0} & \mc{Silver}{0} & \mc{Bronze}{2}\\
Qwen2.5-VL-7B-Instruct &3.5&2.5&5.7&4.4&3.6&7.3&5.5&\cellcolor{Bronze!40}14.7&7.6&9.8&11.5&4.4&3.5&6.5& ~\mc{Gold}{0} & \mc{Silver}{0} & \mc{Bronze}{1}\\
Phi-4-multimodal &2.0&1.6&4.2&3.6&3.6&5.0&4.5&8.3&9.0&10.0&10.1&4.4&5.0&5.5& ~\mc{Gold}{0} & \mc{Silver}{0} & \mc{Bronze}{0}\\
DeepSeek-VL2 &1.8&0.5&2.5&3.4&3.4&5.0&3.2&5.6&4.8&7.3&6.4&5.0&3.9&4.0& ~\mc{Gold}{0} & \mc{Silver}{0} & \mc{Bronze}{0}\\
\bottomrule
\end{tabular}%
}
\caption*{\scriptsize * For \texttt{Kimi-K2-Thinking}, the inference timeout is set to 2 hours, and 9.58\% of cases exceed this limit.}
\vspace{-8mm}
\end{table}

\clearpage

% \clearpage
\section{Discussion}

\subsection{Generalizability of P1-VL}

We conduct specialized post-training to enhance physics problem-solving; however, a critical question remains: \emph{Does this domain-specific optimization compromise or catalyze general STEM reasoning?} 

To investigate the model's generalization capabilities, we evaluate P1-VL on the open-source benchmark \textit{FrontierScience-Olympiad}~\citep{frontierscience}. As detailed in Table~\ref{tab:frontier}, the results demonstrate robust positive transfer: both P1-VL-235B-A22B and P1-VL-30B-A3B achieve significant gains over their base counterparts across all three scientific domains, yielding a total score improvement of 8.0 and 9.1, respectively. Remarkably, even on this text-only benchmark, the multimodal \texttt{P1-VL-235B-A22B} outperforms its text-only sibling (\texttt{P1-235B-A22B}) by a margin of 2.3 points. Furthermore, when augmented with the \textit{PhysicsMinions} agent framework, \texttt{P1-VL-235B-A22B+PhysicsMinions} attains a total score of 67.1, securing state-of-the-art performance among all evaluated open-source models.

\begin{table}[h]\small
\centering
\caption{Evaluation results on the FrontierScience-Olympiad benchmark. The state-of-the-art performance of all the models is marked in bold, and that of open-source models is underlined. $^*$ refers to the evaluation results reported in \cite{frontierscience}.}
\begin{tabular}{lcccc}
\toprule
\textbf{Model} & \textbf{Biology/10} & \textbf{Chemistry/40} & \textbf{Physics/50} & \textbf{Total/100} \\
\midrule
GPT-5.2$^*$ & \textbf{43.5} & \textbf{89.0} & 74.3 & \textbf{77.1} \\
Gemini-3-Pro$^*$ & 41.0 & 85.6 & \textbf{75.5} & 76.1 \\
Claude-Opus-4.5$^*$& 24.0 & 81.8 & 72.5 & 71.4 \\
GPT-5.1$^*$ & 33.5 & 82.4 & 67.4 & 70.0 \\
GPT-5$^*$ & 35.5 & 81.5 & 67.0 & 69.7 \\
\textcolor[HTML]{78206E}{\textbf{P1-VL-235B-A22B+PhysicsMinions}}
	&26.3 &	\underline{77.2} &	67.3 &	\underline{67.1} 	 	 \\
Grok-4$^*$ & 33.0 & 73.2 & 67.2 & 66.2 \\
DeepSeek-V3.2-Thinking & 26.3 & 74.1 & 67.3 & 65.9 \\
\textcolor[HTML]{6c25be}{\textbf{P1-235B-A22B+PhysicsMinions}}  & 30.0 & 71.0 & 68.0 & 65.4 \\
Kimi-K2-Thinking & 20.0 & 76.6 & 65.0 & 65.1 \\
GLM-4.7 & 20.0 & 70.6 & \underline{69.5} & 65.0 \\
\textcolor[HTML]{78206E}{\textbf{P1-VL-235B-A22B}}  & 30.0 & 71.3 & 65.5 & 64.3 \\
% Doubao-seed-1-8-251228-high & 28.8 & 72.8 & 62.8 & 63.4 \\
 o3$^*$ & 30.0 & 76.1 & 59.0 & 62.9 \\
\textcolor[HTML]{6c25be}{\textbf{P1-235B-A22B}}  & 22.5 & 67.2 & 65.8 & 62.0 \\
 o4-mini$^*$ & 41.5 & 71.5 & 57.8 & 61.7 \\
% doubao-seed-1-8-251228-medium & 31.3 & 69.7 & 60.0 & 61.0 \\
GPT-OSS-120B (high) &	\underline{38.8} &	63.4 &	61.5 &	60.0 		\\			
% DeepSeek-V3.2-Chat & 37.5 & 55.9 & 60.8 & 56.5 \\
Qwen3-VL-235B-A22B-Thinking & 26.3 & 61.9 & 57.8 & 56.3 \\
Qwen3-235B-A22B-Thinking-2507 & 26.3 & 58.1 & 57.3 & 54.5 \\
\textcolor[HTML]{6c25be}{\textbf{P1-30B-A3B}} 
 & 15.0 & 61.9 & 56.3 & 54.4 \\
\textcolor[HTML]{78206E}{\textbf{P1-VL-30B-A3B}}  & 20.0 & 58.8 & 54.0 & 52.5 \\
Qwen3-VL-30B-A3B-Thinking	&18.8 &	49.4 &	43.5 &	43.4 \\
Qwen3-30B-A3B-Thinking-2507 & 10.0 & 47.8 & 45.3 & 42.8 \\
o1$^*$ & 20.0 & 50.9 & 40.3 & 42.5 \\
GPT-4o$^*$ & 3.0 & 12.4 & 14.1 & 12.3 \\
\bottomrule
\end{tabular}
\label{tab:frontier}
\end{table}

Besides, we compare the P1-VL series with their respective base models across diverse benchmarks:
% \vspace{-1mm}
\begin{itemize}
    \item {Text Benchmarks}: ten STEM-oriented evaluation datasets (AIME24, AIME25, HMMT-Nov, HMMT-Feb, Brumo, CMICC~\citep{matharena}, IMO-AnswerBench~\citep{imobench}, AMOBench~\citep{an2025amobench}, BeyondAIME~\citep{bytedance_seed_2025_beyondaime}, GPQA~\citep{rein2024gpqa}), and a general reasoning task (LiveBench~\citep{white2024livebench}). 
    \vspace{2mm}
    
    \item Multi-modal Benchmarks: five STEM-oriented benchmarks, including HLE~\citep{phan2025hle}), MMMU~\citep{yue2023mmmu}, MMMU-Pro~\citep{yue2024mmmu}, EMMA-Mini~\citep{hao2025mllmsreasonmultimodalityemma}, and MathVista-Mini \citep{lu2024mathvista}\footnote{For MathVista-Mini and MMMU, we adopt the evaluation scripts posted on the \href{https://github.com/QwenLM/Qwen3-VL/tree/main/evaluation}{Qwen3-VL repository}, and for MMMU-Pro we adopt the setting in \href{https://github.com/MMMU-Benchmark/MMMU/tree/main/mmmu-pro}{MMMU repository}.}.
\vspace{-1mm}
\end{itemize}

\begin{figure}[t]
    \centering
    \includegraphics[width=\linewidth]{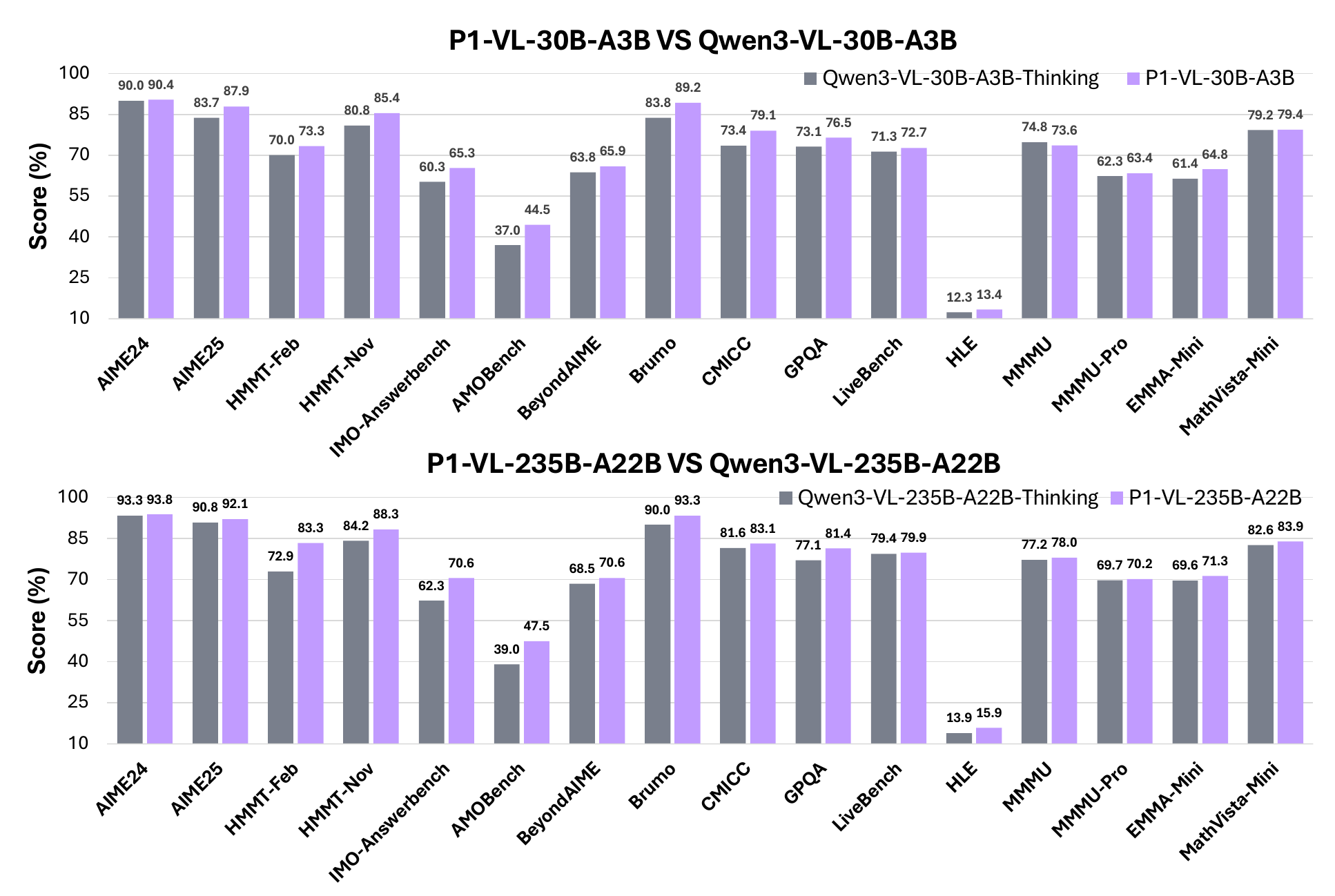}
\caption{Evaluation of P1-VL models against their respective base models on out-of-domain benchmarks. The evaluation suite encompasses both text-only and multi-modal tasks, spanning mathematical reasoning and broader STEM disciplines.}
    \label{fig:general}
\end{figure}
% \vspace{-2mm}

Figure~\ref{fig:general} summarizes the comparative results, illustrating that our models consistently surpass their base counterparts across both text-only and multi-modal benchmarks. For example, in the domain of advanced mathematics, P1-VL-235B-A22B and P1-VL-30B-A3B outperform their respective baselines on AMOBench by margins of 8.5 and 7.5 points. This pattern confirms that  P1-VLs not only retain but actively enhance reasoning capabilities beyond the target physics domain, extending even to rigorous challenges such as IMO-AnswerBench and AMOBench. 
The advantage is not limited to text-based reasoning but extends significantly to visual perception. Taking the multi-image reasoning task EMMA-Mini as a prime example, P1-VL-235B-A22B and P1-VL-30B-A3B achieve a substantial gain of 1.7 and 3.4 points over their baseline. This result specifically highlights the model's enhanced capacity to synthesize information across multiple images and little catastrophic forgetting issue \citep{luo2025empirical} during RL training, validating the effectiveness of our vision-language alignment beyond standard physics problems.

\newpage

\subsection{Train-inference Mismatch}
\label{sec:tim}
\begin{wrapfigure}[19]{r}{0.5\textwidth} % 1. 设置外部容器宽度为页面的一半
    \centering
    % 2. 图片宽度设为容器的 0.95 倍（也就是 \linewidth），这样留有少许边距，防止溢出
    \includegraphics[width=\linewidth]{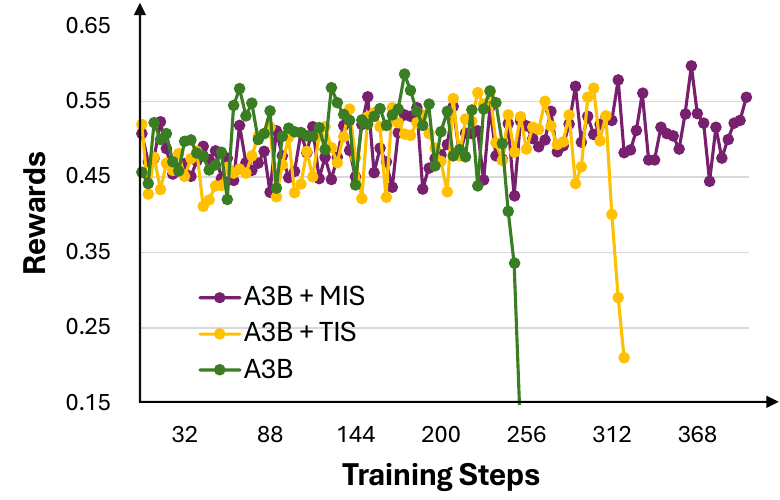}
    \vspace{-11pt} % 3. (可选) 如果下方留白太多，可以用负值的vspace调整
    \caption{Driven by training-inference mismatch, the RL training of Qwen3-VL-30B-A3B-Thinking MoE model suffers from severe collapse. The implementation of Masked Importance Sampling (MIS) proves effective in stabilizing the training. }
    \label{fig:mismatch}
\end{wrapfigure}

During the RL post-training of Qwen3-VL-MoE models on our physics data, we observe a catastrophic training collapse as illustrated in Figure \ref{fig:mismatch}. 
This instability aligns with the {training-inference mismatch} phenomenon characterized in recent analyses~\citep{jiacai2025speed,yao2025tis}, where discrepancies between the inference engine and training framework lead to divergent policy updates. While implementing Truncated Importance Sampling (TIS) \citep{yao2025tis} postpones the onset of this collapse, it still encounters an instability in our experiments. In contrast, adopting {Sequence-Level Masked Importance Sampling (Seq-MIS)} \citep{jiacai2025speed} could stabilize the training process. By strictly rejecting out-of-distribution samples rather than merely clipping their weights, Seq-MIS could maintain more rigorous policy alignment; thus, we adopt Seq-MIS throughout our P1-VL RL post-training.

\subsection{Effect of Training Data Composition}
\label{sec:composition}
\begin{wrapfigure}[13]{r}{0.5\textwidth} % 1. 设置外部容器宽度为页面的一半
    \vspace{-5mm}
    \centering
    \includegraphics[width=.9\linewidth]{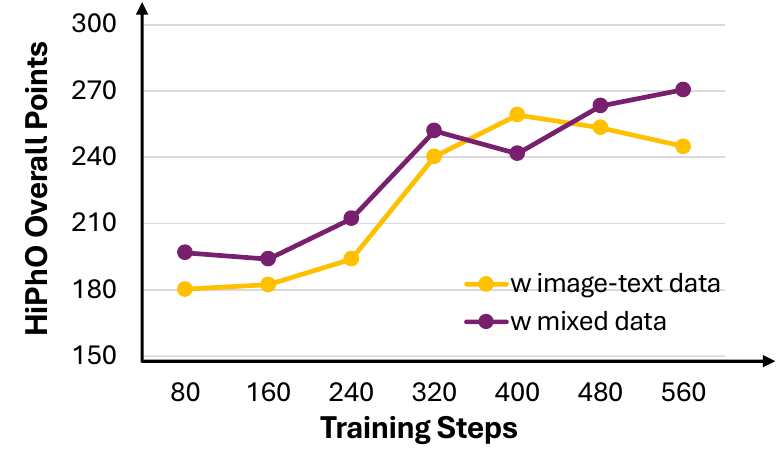}
    % \vspace{-4mm} % 3. (可选) 如果下方留白太多，可以用负值的vspace调整
    \caption{Comparison of Qwen3-VL-4B-Thinking training performance on the image-text data versus the mixed data (text-only + image-text).}
    \label{fig:data_type}
\end{wrapfigure}

We also empirically analyze the impact of training data composition. Figure \ref{fig:data_type} illustrates the performance of Qwen3-VL-4B-Thinking (using XVerifier) when trained exclusively on questions containing images. Crucially, we observe that incorporating text-only data (padded with blank images) induces no negative transfer; rather, it yields superior performance almost all the time. It demonstrates the feasibility and superiority of using mixed multi-modal data. Consequently, we train our P1-VL models using a combination of both the text-only and {image-text data}.

\subsection{Effect of Curriculum Training}\label{sec:curriculum}
\begin{figure}[h]
    \centering
    \includegraphics[width=1\linewidth]{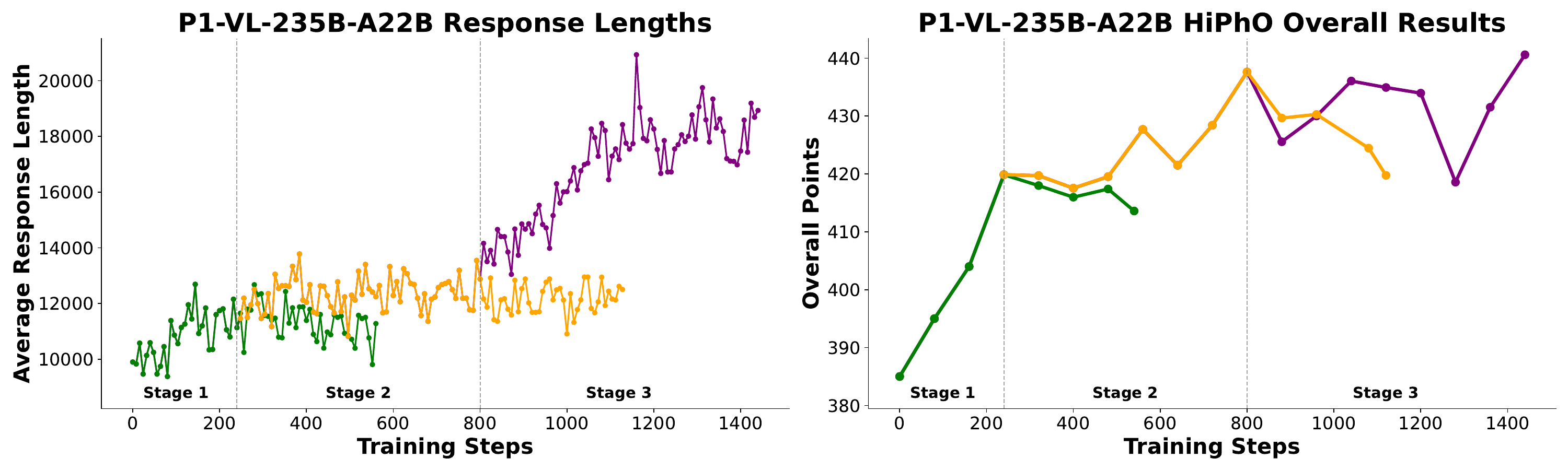}
    \vspace{-3mm}
    \caption{Training dynamics of RL training with and without curriculum difficulty expansion.}
    \label{fig:analysis}
\end{figure}

Figure \ref{fig:analysis} illustrates the impact of curriculum training on model evolution. Each line includes a period of observation after the parameters are fixed.  It shows that in the absence of difficulty expansion, the response length fluctuates stagnantly across Stages 1 and 2, resulting in negligible performance gains. 
Conversely, introducing the curriculum strategy yields a marked increase in both average response length and reasoning accuracy.  These results underscore the critical role of our curriculum RL framework in driving deep reasoning.

% \newpage

\section{Conclusion}
In this work, we presented \textbf{P1-VL}, the first open-source Vision-Language Model family capable of achieving Olympiad-level proficiency in physics. By addressing the critical visual-logical gap inherent in physical problem solving, P1-VL overcomes the limitations of text-only reasoning. We proposed to improve the reasoning capacity by {Curriculum RL Training} to internalize robust reasoning patterns and {Agentic Augmentation} to enable rigorous test-time reflection.
The evaluation results on the HiPhO benchmark, where P1-VL-235B-A22B secured 12 gold medals and demonstrated superior generalization across broader scientific disciplines, underscore the potential of P1-VLs in modeling complex scientific reasoning systems. We posit that the ability to synthesize visual constraints with causal logic is essential for the development of reliable world models and embodied AI. By releasing P1-VL to the research community, we aim to accelerate the transition from symbolic manipulation to genuine machine scientific discovery, paving the way for AI systems that can truly understand and navigate the physical world.

\section{Acknowlegement}
This work is supported by Shanghai AI Laboratory and a locally commissioned task from the Shanghai Municipal Government. 
We would like to extend our special thanks to the developers and maintainers of the following open-source projects, which have been critical to the implementation of this work. This includes Qwen3-VL \citep{bai2025qwen3vltechnicalreport} and Qwen~\citep{yang2025qwen3}, which provided the foundational base models for our research; slime~\citep{slime_github}, whose innovative framework enabled efficient reinforcement learning in our training pipeline; and verl~\citep{sheng2024verl}, which offered a versatile reinforcement learning framework to support model training. We also thank sglang~\citep{zheng2024sglang} for its efficient infrastructure for LLM and VLM serving and inference, and Megatron-LM~\citep{shoeybi2019megatron} for providing the large-scale model training framework.

% \newpage
\bibliography{main}
\bibliographystyle{plainnat}

\newpage
\appendix
\section{Appendix}

\subsection{RL Training on Intern Series}
We further verify the effectiveness of our RL training on the InternVL-3.5-30B-A3B and Intern-S1-mini models. On the HiPhO benchmark, InternVL-3.5-30B-A3B performance improves from 18.3 to 22.6, while Intern-S1-mini sees an increase from 8.5 to 12.4 (utilizing Qwen3-30B-A3B-Instruct as the model-based verifier). Using Gemini-2.5-Flash as the verifier, the trained Intern-S1-mini achieves one silver and ten bronze medals, an improvement over the baseline of ten bronze medals.
These consistent improvements demonstrate that our RL training strategy is effective in eliciting the models' latent capabilities in scientific reasoning.

\subsection{Case Study I: Text+Variable Figure}
\label{sec:case1}
This case study examines a complex problem from the 2025 International Physics Olympiad (IPhO), which investigates the physical principles of Cox's 18th-century timepiece—an ingenious device that harnesses atmospheric pressure fluctuations to generate energy.
The problem requires understanding the figure information to determine the states of the tubes and the behaviors, and analyzing the table information.
The P1-VL-235B-A22B model achieved a perfect score (1.0/1.0) on this problem, demonstrating strong capabilities in:
\begin{itemize}[noitemsep]
    \item \textbf{Physical intuition:} 
    Correctly identifying the critical force balance constraint at the stop position.
   \item \textbf{Visual-Structural Alignment:} Accurately translating the schematic representation of the tube states (liquid levels and gas volumes) into precise hydrostatic equations, effectively bridging visual features with symbolic reasoning.
\item \textbf{Tabular Information Integration:} Synthesizing discrete parameters from the provided table to constrain the variables, specifically extracting the dimensional specifications required to calculate the total work done.
\item \textbf{Long-horizon Logical Reasoning:} Sustaining a coherent derivation chain to link atmospheric pressure fluctuations with mechanical energy storage, correctly applying the Ideal Gas Law without hallucinating non-existent constraints.
\end{itemize}

This performance demonstrates P1-VL's proficiency in tackling competition-level physics problems, requiring a synergy of deep physical understanding, rigorous analytical reasoning, and precise diagrammatic interpretation.

\begin{tcolorbox}[
    enhanced,
    colback=blue!5!white,
    colframe=blue!75!black,
    title=IPhO 2025 Question 2-A.1,
    fonttitle=\bfseries\large,
    breakable
]

\textbf{\large Background:}

In 1765, British clockmaker James Cox invented a clock whose only source of energy is the fluctuations in atmospheric pressure. Cox's clock used two vessels containing mercury. Changes in atmospheric pressure caused mercury to move between the vessels, and the two vessels to move relative to each other. This movement acted as an energy source for the actual clock.

\vspace{0.4cm}

We propose an analysis of this device. Throughout, we assume that 
\begin{itemize}[noitemsep]
\item the Earth's gravitational field $\vec{g} = -g\vec{u_{z}}$ is uniform with $g = 9.8 \mathrm{m} \cdot \mathrm{s}^{-2}$ and $\vec{u_{z}}$ a unit vector; 
\item  all liquids are incompressible and their density is denoted $\\rho$; 
\item  no surface tension effects will be considered; - the variations of atmospheric pressure with altitude are neglected; \item  the surrounding temperature $T_{\mathrm{a}}$ is uniform and all transformations are isothermal.
\end{itemize}

We first consider a bath of water that occupies the semi-infinite space $z \leq 0$. The air above it is at a pressure $P_{\mathrm{a}} = P_{0}$. A cylindrical vertical tube of length $H = 1 \mathrm{m}$, cross-sectional area $S = 10 \mathrm{cm}^{2}$ and mass $m = 0.5 \mathrm{kg}$ is dipped into the bath. The bottom end of the tube is open, and the top end of the tube is closed. We denote $h$ the altitude of the top of the tube and $z_{\ell}$ that of the water inside the tube. The thickness of the tube walls is neglected. 

\vspace{0.4cm}
We start from the situation where the tube in \autoref{fig:placeholder} contains no gas and its top is at the bath level: in other words, \( h = 0 \) and \( z_l = 0 \) (case a). The tube is then slowly lifted until its bottom end reaches the bath level. The pulling force exerted on the tube is denoted \( \vec{F} = F\vec{u}_z \).

\begin{figure}[H]
    \centering
   \includegraphics[width=0.8\linewidth]{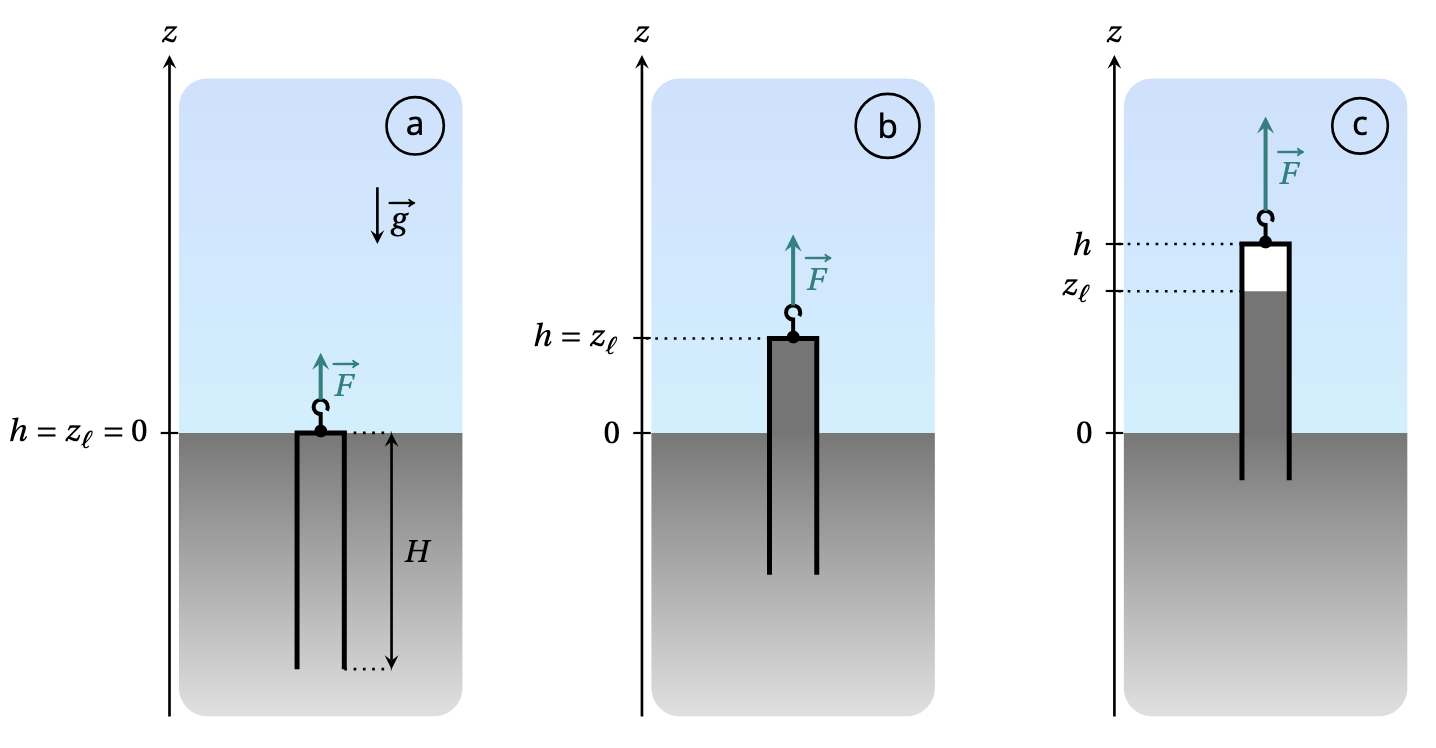}
    \caption{Sketch of the tube in different configurations}
    \label{fig:placeholder}
\end{figure}

\tcblower

\textbf{\large Question:}

For the configuration shown in \autoref{fig:placeholder} (case b), determine the expressions for:
\begin{enumerate}
    \item The pressure $P_w$ in the water at the top of the tube
    
    \item The force $\vec{F}$ necessary to maintain the tube at this position

\end{enumerate}

\vspace{0.2cm}

\textbf{Given data:} $P_0, \rho, m, S, h, g$ and $\vec{u}_z$
\end{tcolorbox}

\begin{tcolorbox}[
    enhanced,
    colback=violet!5!white,
    colframe=violet!75!black,
    title=P1-VL Solution Overview,
    fonttitle=\bfseries,
    breakable
]

To analyze the situation for case b in Cox’s timepiece, we consider the vertical tube of length $ H $, cross-sectional area $ S $, and mass $ m $, lifted so that its bottom end is at the bath surface ($ z = 0 $) and its top is at altitude $ h $. In this configuration, the water level inside the tube is at the same height as the top of the tube, i.e., $ z_{\ell} = h $, indicating that the tube is completely filled with water.

\end{tcolorbox}

\vspace{0.3cm}

\begin{tcolorbox}[
    enhanced,
    colback=violet!5!white,
    colframe=violet!75!black,
    title={(A.1) $ P_{\rm w} $ and $\vec{F}$},
    fonttitle=\bfseries,
    breakable
]
The tube is sealed at the top and open at the bottom, and the water inside is in hydrostatic equilibrium. The pressure at the bottom of the tube (at $ z = 0 $) must equal the atmospheric pressure $ P_0 $, as the tube is open to the bath at this level.

Using the hydrostatic pressure law, the pressure at any height $ z $ inside the water column is given by:
\begin{equation}
P(z) = P_0 - \rho g z
\end{equation}
At the top of the tube, where $ z = h $, the pressure in the water is:
\begin{equation}
P_{\rm w} = P_0 - \rho g h
\end{equation}This result reflects that the pressure at the top of the water column is reduced by the weight of the water column of height $ h $.

%\tcblower
%\textbf{Answer:} \quad $\boxed{P_{\rm w} = P_0 - \rho g h}$

%\end{tcolorbox}

\vspace{1cm}

To determine the force $ \vec{F} $ necessary to hold the tube in place, we perform a force balance on the tube alone.

\textbf{Forces on the Tube:}

\begin{itemize} 
\item \textbf{Gravitational force:} The tube has a mass $ m $, so its weight is $ -mg\vec{u_z} $.
\item \textbf{Pressure forces:} The atmospheric pressure $ P_0 $ acts downward on the top cap of the tube, exerting a force $ -P_0 S \vec{u_z} $.The water pressure $ P_{\rm w} $ at the top of the tube acts upward on the inner surface of the cap, exerting a force $ +P_{\rm w} S \vec{u_z} $.
\item \textbf{Applied force:} $ \vec{F} $ is applied upward to maintain equilibrium.
\end{itemize}

Thus, the \textbf{net pressure force} on the tube is:
\begin{equation}
(P_{\rm w} - P_0) S \vec{u_z} = -\rho g h S \vec{u_z}
\end{equation}
This is a downward force due to the pressure difference between the outside atmosphere and the water column inside the tube.

The net force on the tube must be zero:
\begin{equation}
\vec{F} + (P_{\rm w} - P_0) S \vec{u_z} - m g \vec{u_z} = \vec{0}
\end{equation}
Substituting $ P_{\rm w} = P_0 - \rho g h $:
\begin{equation}
\vec{F} - \rho g h S \vec{u_z} - m g \vec{u_z} = \vec{0} \Rightarrow \vec{F} = (m g + \rho g h S) \vec{u_z}
\end{equation}
This result is physically intuitive: the force required to hold the tube in place must counteract both the tube's own weight and the net downward pressure force due to the difference between atmospheric pressure and the reduced pressure at the top of the water column.

\tcblower
\textbf{Answer:} \quad $\boxed{P_{\rm w} = P_0 - \rho g h}$ \quad $\boxed{\vec{F} = (m g + \rho g h S) \vec{u_z}}$
\end{tcolorbox}

\begin{tcolorbox}[
    enhanced,
    colback=blue!5!white,
    colframe=blue!75!black,
    title=IPhO 2025 Question 2-A.2,
    fonttitle=\bfseries\large,
    breakable
]

\textbf{\large Background:}
Three experiments are performed. In each, the tube is lifted from the initial state shown in \autoref{fig:placeholder}(a)
under the conditions specified in \autoref{fig:Table1}.

\begin{figure}[H]
    \centering
    \includegraphics[width=0.6\linewidth]{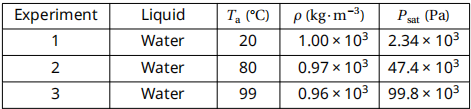}
    \caption{Experimental conditions and numerical values of physical quantities for each experiment
($P_{\rm sat}$ designates the saturated vapour pressure of the pure fluid)}
    \label{fig:Table1}
\end{figure}

In each case, we study the evolution of the force $F$ that must be applied in order to maintain the tube in equilibrium at an altitude $h$, the external pressure being fixed at $P_{\mathrm{a}} = P_0 = 1.000 \times 10^5 \, \mathrm{Pa}$. Two different behaviours are possible

\begin{figure}[H]
    \centering
    \includegraphics[width=0.8 \linewidth]{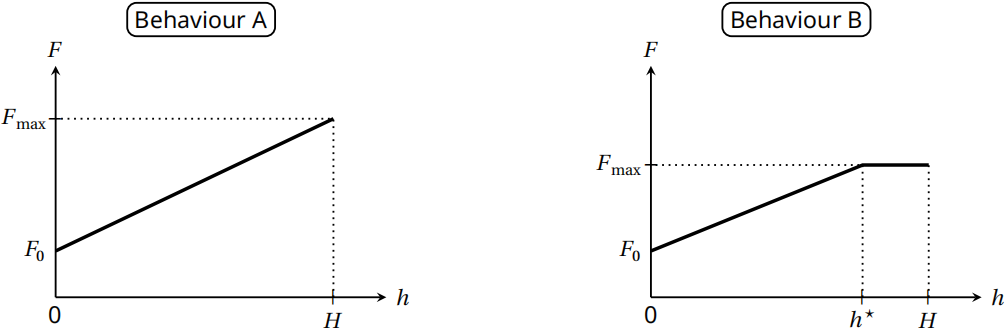}
    \caption{Two different behaviours}
    \label{fig_behaviours}
\end{figure}
\vspace{0.4cm}

\tcblower

\textbf{\large Question:}

For each experiment, complete the table in the answer sheet to indicate the expected behaviour and the numerical values for $F_{\mathrm{max}}$ and for $h^*$ (when pertinent), where $F_{\mathrm{max}}$ and $h^*$ are defined in the figures illustrating the two behaviours.

\vspace{0.2cm}

\end{tcolorbox}

\begin{tcolorbox}[
    enhanced,
    colback=violet!5!white,
    colframe=violet!75!black,
    title=P1-VL Solution Overview,
    fonttitle=\bfseries,
    breakable
]

To determine the behavior of the tube during lifting and the corresponding values of $ F_{\max} $ and $ h^* $, we analyze the pressure equilibrium and the conditions under which vaporization (cavitation) occurs.

\end{tcolorbox}

\vspace{0.3cm}

\begin{tcolorbox}[
    enhanced,
    colback=violet!5!white,
    colframe=violet!75!black,
    title={(A.2) Answer Sheet},
    fonttitle=\bfseries,
    breakable
]

The vertical cylinder, sealed at the top and open at the bottom, is lifted from the bath (where $h = 0$) to the position where its bottom end reaches the surface (i.e., $h = H = 1 \, \mathrm{m}$). The pressure at the top of the tube is governed by:
\begin{equation}
P_{\mathrm{top}} = P_0 - \rho g h
\end{equation}

Cavitation (formation of vapor) occurs when $P_{\mathrm{top}} = P_{\mathrm{sat}}$, i.e., when:
\begin{equation}
    h^* = \frac{P_0 - P_{\mathrm{sat}}}{\rho g}
\end{equation}

\begin{itemize}
    \item \textbf{Behavior A} occurs when $h^* > H$, meaning the tube remains completely filled with liquid throughout the lifting process ($z_\ell = h$).
    \item \textbf{Behavior B} occurs when $h^* < H$, meaning the water column inside the tube stops rising at $h^*$, and the top of the tube continues to rise while the water level inside remains at $z_\ell = h^*$.
\end{itemize}

The force $F$ required to hold the tube in equilibrium is given by:
\begin{equation}
    F = mg + (P_0 - P_{\mathrm{top}})S
\end{equation}

\begin{itemize}
    \item For \textbf{Behavior A}, $P_{\mathrm{top}} = P_0 - \rho g h$, so:
    \begin{equation}
    F = mg + \rho g S h
    \end{equation}
    At $h = H$, 
    \begin{equation}
    F_{\max} = mg + \rho g S H
    \end{equation}
    \item For \textbf{Behavior B}, once $h > h^*$, $P_{\mathrm{top}} = P_{\mathrm{sat}}$, so:
    \begin{equation}
    F_{\max} = mg + (P_0 - P_{\mathrm{sat}})S
    \end{equation}
\end{itemize}

\textbf{Data Recap}
\begin{itemize}
    \item Cross-sectional area: $S = 10 \, \mathrm{cm^2} = 1 \times 10^{-3} \, \mathrm{m^2}$
    \item Mass of the tube: $m = 0.5 \, \mathrm{kg}$
    \item Gravitational acceleration: $g = 9.8 \, \mathrm{m/s^2}$
    \item Atmospheric pressure: $P_0 = 1.000 \times 10^5 \, \mathrm{Pa}$
\end{itemize}

\textbf{Experiment 1: Water at $20^\circ \mathrm{C}$}
\begin{itemize}
    \item $\rho = 1.00 \times 10^3 \, \mathrm{kg/m^3}$
    \item $P_{\mathrm{sat}} = 2.34 \times 10^3 \, \mathrm{Pa}$
\end{itemize}
\begin{equation}
h^* = \frac{1.000 \times 10^5 - 2.34 \times 10^3}{1.00 \times 10^3 \times 9.8} \approx 9.96 \, \mathrm{m} > H    
\end{equation}

$\rightarrow$ \textbf{Behavior A}
\begin{equation}
F_{\max} = mg + \rho g S H = 0.5 \times 9.8 + 1.00 \times 10^3 \times 9.8 \times 1 \times 10^{-3} = 4.9 + 9.8 = 14.7 \, \mathrm{N}
\end{equation}

\textbf{Experiment 2: Water at $80^\circ \mathrm{C}$}
\begin{itemize}
    \item $\rho = 0.97 \times 10^3 \, \mathrm{kg/m^3}$
    \item $P_{\mathrm{sat}} = 47.4 \times 10^3 \, \mathrm{Pa}$
\end{itemize}
\begin{equation}
h^* = \frac{1.000 \times 10^5 - 47.4 \times 10^3}{0.97 \times 10^3 \times 9.8} \approx 5.53 \, \mathrm{m} > H
\end{equation}
$\rightarrow$ \textbf{Behavior A}
\begin{equation}
F_{\max} = mg + \rho g S H = 4.9 + 0.97 \times 10^3 \times 9.8 \times 1 \times 10^{-3} \approx 4.9 + 9.506 = 14.4 \, \mathrm{N}
\end{equation}

\textbf{Experiment 3: Water at $99^\circ \mathrm{C}$}
\begin{itemize}
    \item $\rho = 0.96 \times 10^3 \, \mathrm{kg/m^3}$
    \item $P_{\mathrm{sat}} = 99.8 \times 10^3 \, \mathrm{Pa}$
\end{itemize}
\begin{equation}
h^* = \frac{1.000 \times 10^5 - 99.8 \times 10^3}{0.96 \times 10^3 \times 9.8} \approx 0.0213 \, \mathrm{m} = 2.1 \, \mathrm{cm} < H
\end{equation}
$\rightarrow$ \textbf{Behavior B}
\begin{equation}
F_{\max} = mg + (P_0 - P_{\mathrm{sat}})S = 4.9 + (200)(1 \times 10^{-3}) = 4.9 + 0.2 = 5.1 \, \mathrm{N}
\end{equation}

\tcblower
\textbf{Answer:} 

\begin{center}
\begin{tabular}{|c|c|c|c|}
\hline
Experiment & Behaviour (A or B?) & $h^*$ (cm) & $F_{\max}$ (N) \\
\hline
1 & A & None & 14.7 \\
2 & A & None & 14.4 \\
3 & B & 2.1 & 5.1 \\
\hline
\end{tabular}
\end{center}
\vspace{0.1cm}
\end{tcolorbox} 

\end{document}